\documentclass{article}

\usepackage[preprint]{neurips_2025}

\usepackage[utf8]{inputenc} 
\usepackage[T1]{fontenc}    
\usepackage{hyperref}       
\usepackage{url}            
\usepackage{booktabs}       
\usepackage{amsfonts}       
\usepackage{nicefrac}       
\usepackage{microtype}      
\usepackage{xcolor}         
\usepackage{amsmath}
\usepackage{graphicx}
\usepackage{wrapfig}
\usepackage{caption}
\usepackage{multirow}
\usepackage{float}
\usepackage{tabularx}

\usepackage{amsmath}
\usepackage{amssymb}
\usepackage{mathtools}
\usepackage{amsthm}

\title{GeneMamba: An Efficient and Effective Foundation Model on Single Cell Data}

\author{%
  Cong Qi\\
  Department of Computer Science\\
  New Jersey Institute of Technology\\
  \And
  Hanzhang Fang\\
  Department of Computer Science\\
  New Jersey Institute of Technology\\
  \AND
  Siqi Jiang\\
  Department of Computer Science\\
  New Jersey Institute of Technology\\
  \AND
  Xun Song\\
  Department of Computer Science\\
  New Jersey Institute of Technology\\
  \And
  Tianxing Hu\\
  Department of Computer Science\\
  New Jersey Institute of Technology\\
  \And
  Wei Zhi\\
  Department of Computer Science\\
  New Jersey Institute of Technology\\
}

\begin{document}

\maketitle

\begin{abstract}
Transformer-based architectures have shown promise across various domains but struggle with computational inefficiency and scalability. To address these challenges, we introduce GeneMamba, a novel model specifically designed for single-cell data analysis. GeneMamba incorporates the BiMamba module to efficiently capture gene context information and employs biologically informed loss functions during training. The model enables scalable processing of over 50 million cells while significantly reducing computational costs. It delivers strong performance in multi-batch integration, cell type annotation, and gene pair correlation analysis. Furthermore, reconstruction experiments highlight GeneMamba’s explainability, establishing it as a robust foundation for advancing single-cell transcriptomics in biological and biomedical research. By unifying bidirectional state space dynamics with rank-based gene tokenization, GeneMamba achieves both contextual depth and biological interpretability. These capabilities position GeneMamba as a scalable and principled backbone for future foundation models in single-cell omics.
\end{abstract}

\section{Introduction}

Single-cell RNA sequencing (scRNA-seq) has fundamentally transformed our ability to study cellular heterogeneity and dynamics by enabling high-resolution profiling of gene expression at the individual cell level \cite{korsunsky2019fast, theodoris2023transfer}. This technology allows researchers to dissect complex cellular compositions within tissues, trace differentiation trajectories, and identify rare cell populations that remain undetectable with bulk RNA sequencing \cite{shen2023generative, wen2023single, deconinck2021recent}. The rich transcriptomic insights provided by scRNA-seq have catalyzed significant advancements in developmental biology, disease modeling, and drug discovery \cite{gu2023mamba, yang2024genecompass, li2024single}.

Despite these advantages, the computational analysis of scRNA-seq data presents formidable challenges due to its high dimensionality, inherent sparsity, and technical noise \cite{du2019gene2vec, zhao2023large}. Addressing these challenges requires robust computational models capable of capturing biologically meaningful patterns while efficiently handling the scale and variability of single-cell data. Transformer-based architectures, such as scBERT \cite{yang2022scbert} and scGPT \cite{cui2024scgpt}, have emerged as powerful tools in this domain, demonstrating strong performance in tasks such as cell type classification, gene expression imputation, and differential expression analysis \cite{bian2024scmulan, wen2023cellplm, szalata2024transformers}.

However, transformers exhibit critical limitations when applied to scRNA-seq data. The quadratic complexity of their self-attention mechanism constrains their scalability for long sequences, which are characteristic of single-cell transcriptomes \cite{vaswani2017attention, choromanski2020rethinking}. Moreover, transformers often struggle to effectively capture long-range dependencies, which are essential for modeling gene regulatory interactions and cell state transitions \cite{dao2022flashattention, chen2023transformer}. These limitations have driven the exploration of alternative architectures, among which state space models (SSMs) have emerged as promising solutions, offering improved efficiency and scalability for processing long sequences \cite{gu2023mamba, mamba2}.

In this study, we introduce \textbf{GeneMamba}, a novel state space model designed to efficiently train large-scale cell models on scRNA-seq data. SSM-based models have demonstrated competitive or superior performance compared to transformers while significantly reducing computational overhead~\cite{shen2022universal, gong2024xtrimogene, ryu2023integration}. Building upon this foundation, GeneMamba incorporates bidirectional computation, enhancing its ability to capture both upstream and downstream contextual dependencies, thereby improving its performance in tasks requiring global contextual awareness \cite{liu2024bidirectional, liang2024bi}. We validate GeneMamba across a diverse set of applications, including multi-batch integration, cell type annotation, and gene-gene correlation analysis. Our experimental results demonstrate substantial improvements in computational efficiency and predictive performance compared to existing models \cite{hao2024large, sun2024bidirectional, petegrosso2020machine}. Furthermore, to assess its scalability, we conduct a comprehensive performance analysis on various variations of GeneMamba (Appendix~\ref{section:scalability}), highlighting its advantages in real-world biological applications. The code is available at https://github.com/MineSelf2016/GeneMamba. The HuggingFace usage guide is at https://huggingface.co/mineself2016/GeneMamba.

Our contributions are threefold:

1. We present GeneMamba, a scalable and efficient model designed for single-cell RNA sequencing data, which harnesses the strengths of the SSM architecture. GeneMamba's architecture is tailored to tackle the challenges of high-dimensional and sparse scRNA-seq data, offering a robust and flexible framework for single-cell analysis.

2. We validate the effectiveness of GeneMamba in multiple downstream tasks, showcasing its potential to advance single-cell transcriptomics research. Extensive experiments highlight GeneMamba's versatility and robustness, making it a valuable tool for the single-cell research community.

3. We demonstrate GeneMamba's superior reconstruction ability compared to transformer-based models, providing insights into the interpretability and effectiveness of state space models.

\section{Related Work}

\subsection{Discretization of Gene Expression}

Discretizing gene expression levels into tokens is a crucial step in single-cell transcriptomics modeling. Existing methods employ various tokenization strategies, each with distinct advantages and limitations.

Bin-based discretization, as used by scBERT \cite{yang2022scbert}, scGPT \cite{cui2024scgpt}, and scMulan \cite{bian2024scmulan}, groups expression values into predefined bins. This approach preserves absolute value distributions and simplifies sequence modeling, but may introduce information loss, particularly for genes with subtle but biologically significant expression differences. Additionally, binning can be sensitive to parameter selection, affecting downstream results.

Value projection\cite{szalata2024transformers}, adopted by scFoundation \cite{hao2024large} and its backbone model xTrimoGene \cite{gong2024xtrimogene}, projects gene expression values into continuous embeddings rather than discrete categories. This method maintains full data resolution by applying a linear transformation to the gene expression vector, which is then combined with gene-specific embeddings. However, the use of continuous embeddings diverges from traditional tokenization strategies in NLP-based transformers, and its impact on model performance remains an open question.

Rank-based discretization, utilized by Geneformer \cite{theodoris2023transfer}, GeneCompass \cite{yang2024genecompass}, tGPT\cite{shen2023generative} and LangCell \cite{zhao2024langcell}, transforms gene expression values into ordinal rankings. This approach effectively captures relative expression levels and is more robust to batch effects and noise. Our method is based on rank-based discretization, as employed in Geneformer, which aligns naturally with biological processes such as regulatory interactions.

\subsection{Model Architectures for Single-Cell Analysis}

Transformer-based architectures, used by scBERT, scGPT, Geneformer, and scFoundation, have been successfully applied to single-cell data, leveraging the power of deep learning to model complex gene expression patterns. These models have shown promising results in various tasks, including cell-type annotation, batch integration, and multiomics analysis. However, transformer\cite{vaswani2017attention} has inherent limitations. The most significant one is their quadratic computational complexity with respect to sequence length, which makes them less feasible for long sequences typical of scRNA-seq data. This issue arises because the self-attention mechanism used in transformer requires computing attention scores for all pairs of tokens, leading to inefficiencies in handling long sequences. Additionally, transformer often struggles with capturing long-range dependencies in sequences, which is crucial for understanding gene regulatory networks and cell state transitions. Mamba\cite{gu2023mamba} has been introduced to address the transformer efficiency problem. Bidirectional Mamba(Bi-Mamba) has recently been used to solve long-sequence problems in tasks such as feature extraction\cite{sun2024bidirectional}, sequential recommendation\cite{liu2024bidirectional}, and time series forecasting\cite{liang2024bi}. Bi-Mamba is designed to efficiently process ultra-long sequences with linear computational complexity, offering a significant reduction in memory and computation requirements compared to transformer. By leveraging state-space dynamics, Bi-Mamba can capture long-range dependencies effectively, making it well-suited for modeling gene regulatory interactions and cell state transitions. Furthermore, Bi-Mamba's bidirectional processing enables the simultaneous consideration of upstream and downstream contexts, enhancing its ability to model complex dependencies in single-cell data. This architecture represents a promising alternative to transformer-based methods, addressing their key limitations while maintaining high performance in various single-cell analysis tasks.

More related work is provided in Appendix~\ref{section:more_related_work}.

\section{Methods}

\begin{figure*}[ht]
\begin{center}
\centerline{\includegraphics[width=\textwidth]{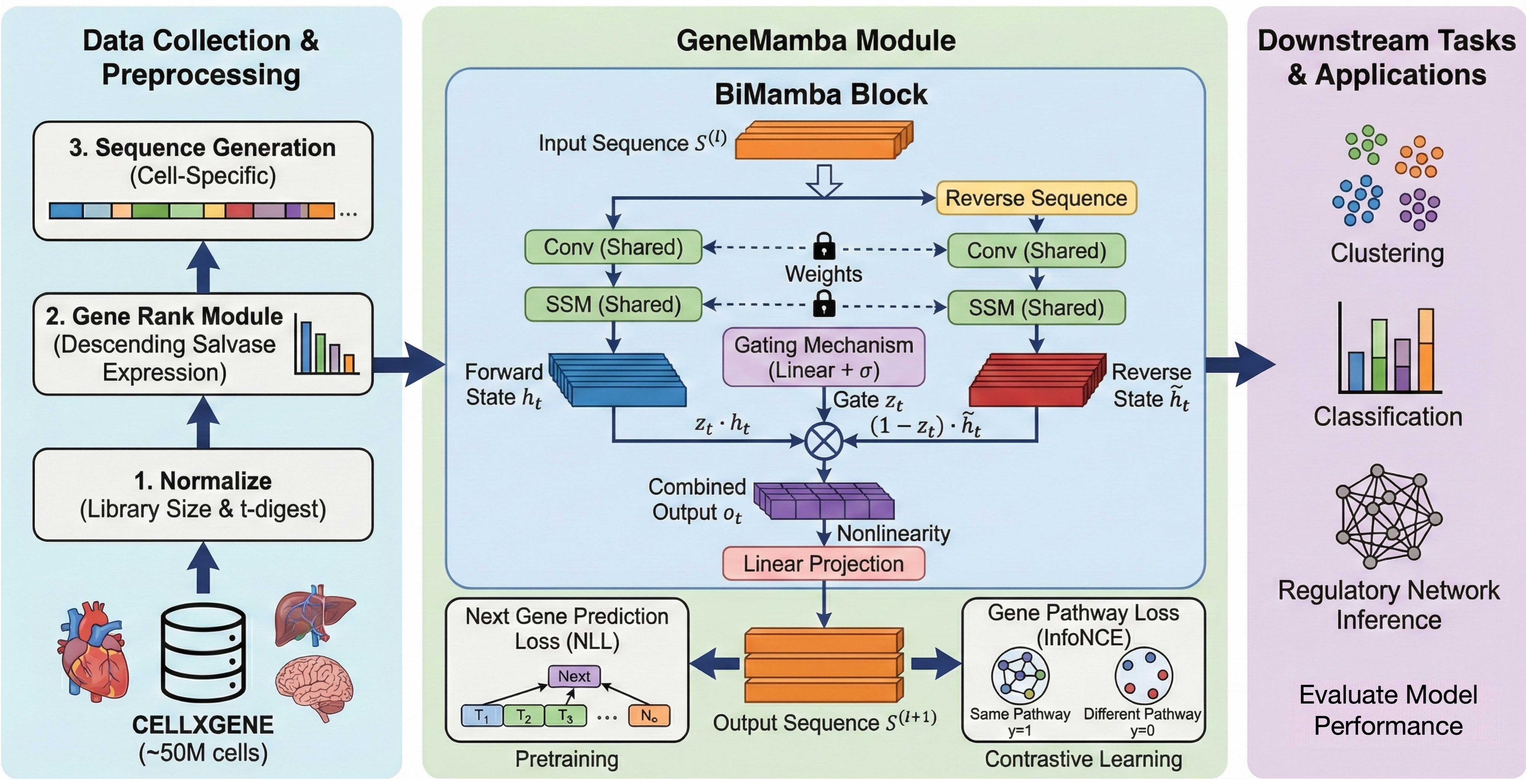}}
\caption{The GeneMamba architecture and its downstream task applications. The framework begins with the collection of training data (approximately 50M cells) from CELLXGENE, encompassing a diverse array of tissues and organs. After preprocessing, the data is prepared through a Gene Rank Module to transform single-cell data into input sequences. The GeneMamba module then captures the contextual information within each single cell. Once pretrained, the model and its embeddings are employed for various downstream tasks to evaluate the model's performance.}
\label{fig:framework_overview}
\end{center}
\end{figure*}

\begin{wrapfigure}{r}{0.5\textwidth}
\centering
\includegraphics[width=1.0\linewidth]{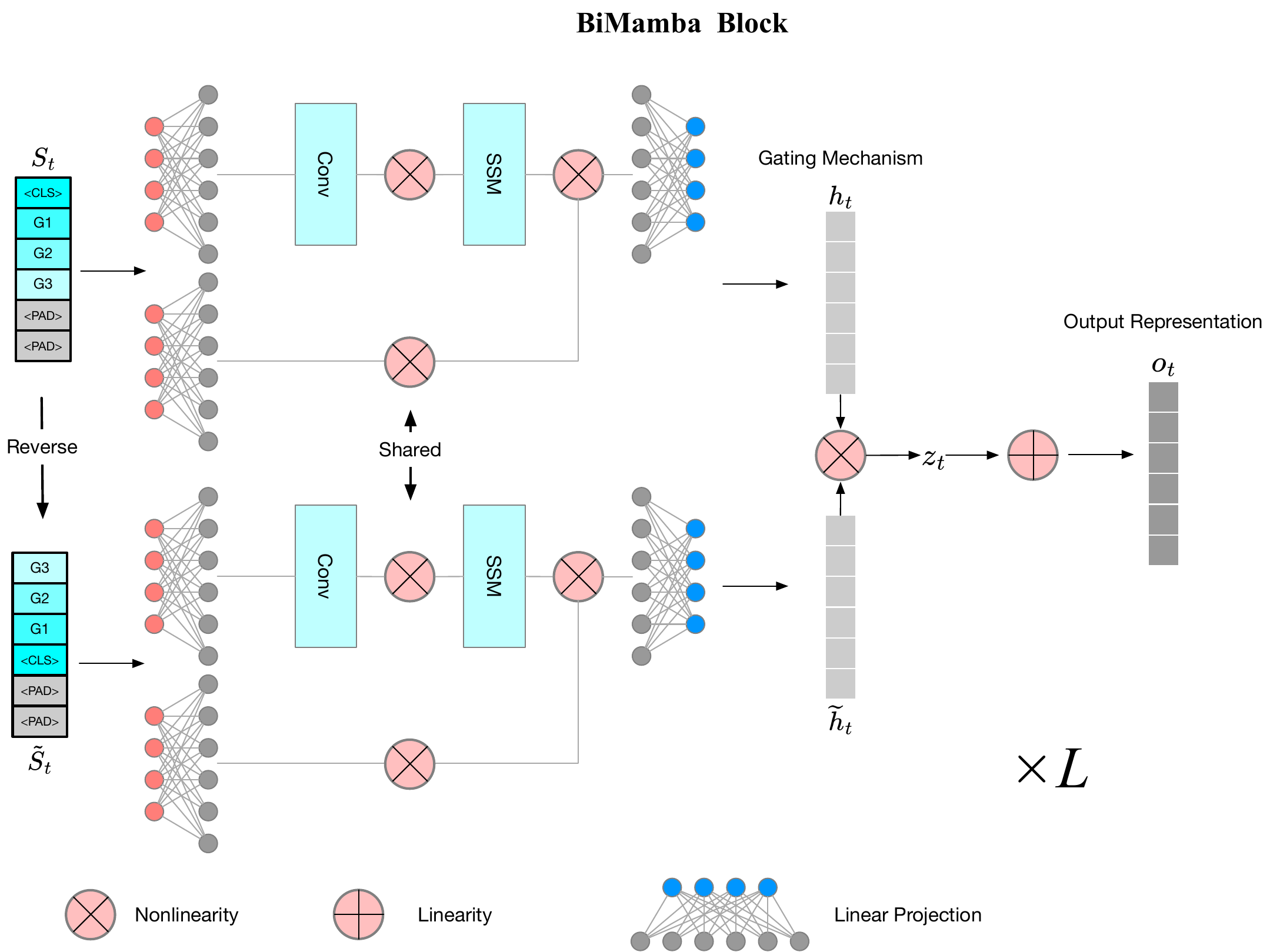}
\caption{The schematic overview of BiMamba Block. The BiMamba Block processes input sequences bidirectionally, capturing forward and reverse context through shared convolutional layers (Conv) and structured state machines (SSM). A gating mechanism integrates the outputs, followed by linear projection and nonlinearity layers, generating a context-aware representation for downstream tasks.}
\label{fig:framework_bimamba}
\end{wrapfigure}

In this section, we introduce the framework of GeneMamba (Figure~\ref{fig:framework_overview}), the designed BiMamba module (Figure~\ref{fig:framework_bimamba}), and the pretraining objective (Section~\ref{section:pretraining objective}).

\subsection{Data Processing}
A significant challenge in modeling single-cell data is that the input data is not a plain token sequence; instead, it comprises both gene tokens and their corresponding expression values. To address this, we represent the input data as a gene expression matrix \( M \in \mathbb{R}^{c \times g} \), where \( c \) is the number of cells, and \( g \) is the number of genes. In this matrix, rows correspond to cells, columns correspond to genes, and unexpressed genes have a zero expression value.

First, we preprocess the single-cell data using standard techniques (Appendix~\ref{section:data_curation}). Next, we normalize the matrix to account for sequencing depth and gene-specific variation. Specifically, each element \( M_{ij} \) (the expression value of gene \( j \) in cell \( i \)) is first divided by the total expression of all genes in cell \( i \). Then, we compute the median of the non-zero expression values for each gene \( j \) across all cells using the t-digest algorithm for efficient computation. The final normalized expression value is given by:
\begin{equation}
M_{ij}^{\text{norm}} = \frac{M_{ij} / \sum_{k=1}^n M_{ik}}{\text{t-digest}\{M_{kj} \mid M_{kj} > 0\}}
\end{equation}

Finally, we rank the genes within each cell in descending order based on their normalized expression values. This ranking approach highlights genes that distinguish cell states while deprioritizing universally high-expression housekeeping genes, ensuring they are assigned lower ranks in downstream analyses.
\begin{equation}
R_i=\operatorname{argsort}\left(-M_{ij}^{\text{norm}}\right), \quad \forall j    
\end{equation}

\subsection{Bi-Mamba Architecture}

State Space Models (SSMs) provide a powerful framework for modeling sequences by utilizing a latent state that evolves over time. At each time step, the latent state \( h_t \) is updated based on the previous state \( h_{t-1} \), the current input \( x_t \), and the system’s parameters. The dynamics of the system are described as:
\begin{equation}
h_t = A h_{t-1} + B x_t, \quad y_t = C h_t    
\end{equation}

where \( A, B, \) and \( C \) are matrices defining how the input and state interact. This formulation enables the model to represent long-range dependencies in sequences efficiently. The convolutional view reformulates the output \( y \) as:

\begin{equation}
y = x * K, K = [CB, CAB, CA^2B, \dots]    
\end{equation}

where \( K \) is a sequence-length-dependent kernel capturing temporal relationships.

However, traditional SSMs are limited by their static nature; parameters like \( A, B, C \) are constant, making them unable to adapt dynamically to the input content. To address this, the Mamba model introduces input-dependent dynamics.

In Mamba, the parameters \( A, B, \Delta \), and \( C \) are functions of the input, allowing the system to dynamically adjust to the sequence content. The updated equations are:

\begin{equation}
h_t = f_A(x_t) h_{t-1} + f_B(x_t) x_t, \quad y_t = f_C(x_t) h_t    
\end{equation}

where \( f_A, f_B, \) and \( f_C \) are learned transformations of the input \( x_t \). By making these parameters input-dependent, Mamba enables selective propagation of relevant information while filtering out noise, significantly enhancing the model’s ability to capture complex sequence dynamics. Mamba is also computationally efficient, achieving linear scaling with sequence length.

While Mamba processes sequences in a unidirectional manner, many tasks require bidirectional context to fully capture dependencies. To address this limitation, we extend Mamba to a bidirectional version, called \textbf{Bi-Mamba}.

The Bi-Mamba model processes the input sequence in both forward and reverse directions to capture bidirectional contextual relationships. Given an input sequence \( S = [s_1, s_2, \dots, s_n] \), the following steps outline the processing pipeline in Bi-Mamba:

1. \textbf{Reversing the Input}: A reversed version of the sequence, \( S_{\text{rev}} = [s_n, s_{n-1}, \dots, s_1] \), is created. During the reversing and combining process, padding tokens are handled separately to prevent alignment artifacts. Only valid tokens (e.g., sequence content) are flipped, while padding remains static. This ensures that information flows in both directions during processing.

2. \textbf{Parallel Processing}: Both the original sequence \( S \) and the reversed sequence \( S_{\text{rev}} \) are processed independently using identical Mamba layers with shared weights. For each layer \( l \), the outputs are:
\begin{equation}
h_t^{(l)} = f_A^{(l)}(s_t) h_{t-1}^{(l)} + f_B^{(l)}(s_t) s_t    
\end{equation}

for the original sequence, and:
\begin{equation}
\tilde{h}_t^{(l)} = f_A^{(l)}(s_t) \tilde{h}_{t-1}^{(l)} + f_B^{(l)}(s_t) s_t    
\end{equation}

for the reversed sequence \( S_{\text{rev}} \), where \( \tilde{h}_t^{(l)} \) represents the reversed latent state.

3. \textbf{Combining the Outputs}: The outputs from the forward and reversed passes are combined to form a unified representation. Instead of summing the outputs directly, Bi-Mamba introduces a learnable gating mechanism to balance the contributions of forward and backward information:

\begin{equation}
z_t^{(l)} = \sigma(W^{(l)} [h_t^{(l)}, \tilde{h}_t^{(l)}])    
\end{equation}

\begin{equation}
o_t^{(l)} = z_t^{(l)} \cdot h_t^{(l)} + (1 - z_t^{(l)}) \cdot \tilde{h}_t^{(l)}    
\end{equation}
   where \( \sigma \) is a sigmoid function, \( W^{(l)} \) is a learnable weight matrix, and \( o_t^{(l)} \) is the combined output for token \( t \) at layer \( l \).

4. \textbf{Stacking Layers}: The Bi-Mamba process is repeated across multiple layers, with each layer refining the bidirectional representations:

\begin{equation}
S^{(l+1)} = \text{Bi-Mamba}(S^{(l)}), \quad S^{(0)} = S    
\end{equation}

The final sequence representation \( S^{(L)} \) encodes rich bidirectional dependencies, making it suitable for tasks requiring global context.

The Bi-Mamba architecture allows us to capture bidirectional contextual relationships by processing sequences in both forward and reverse directions. With the gating mechanism, we can seamlessly integrate these contexts, enabling meaningful applications like cell type annotation and gene interaction prediction. By leveraging a shared-weight design and linear scalability, we enhance the strengths of Mamba, creating a versatile and efficient tool for sequence modeling that aligns with our goals.

\subsection{Pretraining Objective}
\label{section:pretraining objective}

\textbf{Next Gene Prediction Loss}
The sequence modeling module processes gene expression sequences to predict the probability distribution of the next gene token conditioned on all previous gene tokens within a cell. The loss function, referred to as \(\mathcal{L}_{\text{lang}}\), is computed as the negative log-likelihood (NLL) of the true next gene token \(g_j\) given the preceding tokens \(\{g_1, g_2, \dots, g_{j-1}\}\), represented as:

\begin{equation}
\mathcal{L}_{\text{lang}} = -\frac{1}{M} \sum_{j=1}^M \log P(g_j \mid g_1, g_2, \dots, g_{j-1})    
\end{equation}

This loss function enables end-to-end training of the model by leveraging the sequential representation of gene expression data to predict each gene based on its preceding context within the same cell. By stacking multiple BiMamba layers, the model is trained to capture complex bidirectional relationships among genes, enabling robust analysis and modeling of single-cell gene expression data.

\textbf{Gene Pathway Loss}

To capture the gene-gene relationships within biological pathways, we employ an InfoNCE loss function that enforces similarity among genes sharing a common pathway~\cite{oord2018representation, sharma2023regenne, gundogdu2022integrating}. Pathways represent functional groupings of genes that collectively contribute to specific biological processes~\cite{zien2000analysis, garcia2015pathway}. Genes within the same pathway are often functionally related and co-regulated, making their representations in the embedding space naturally similar. Conversely, genes that belong to different pathways should be distinct in their embeddings. To achieve this, we define gene pairs within the same pathway as positive pairs (label \(y_{ij} = 1\)) and gene pairs from different pathways as negative pairs (label \(y_{ij} = 0\)).

Given a gene pair \((i, j)\), we compute the cosine similarity between their embeddings:

\begin{equation}
\text{sim}(i, j) = \frac{\mathbf{h}_i \cdot \mathbf{h}_j}{\|\mathbf{h}_i\| \|\mathbf{h}_j\|}    
\end{equation}

where \(\mathbf{h}_i\) and \(\mathbf{h}_j\) are the normalized embeddings of genes \(i\) and \(j\). Positive pairs correspond to genes sharing a pathway (label \(y_{ij} = 1\)), while all other gene pairs in the batch are treated as negatives (label \(y_{ij} = 0\)). To emphasize the contrast between positive and negative pairs, we normalize the similarities using temperature scaling:

\begin{equation}
\tilde{\text{sim}}(i, j) = \frac{\text{sim}(i, j)}{\tau}    
\end{equation}

where \(\tau > 0\) is a temperature hyperparameter. The InfoNCE loss for a batch of gene pairs is then formulated as:

\begin{equation}
\mathcal{L}_{\text{pathway}} = - \frac{1}{|\mathcal{P}|} \sum_{(i, j) \in \mathcal{P}} \log \frac{\exp(\tilde{\text{sim}}(i, j))}{\sum_{k \in \mathcal{N}(i)} \exp(\tilde{\text{sim}}(i, k))}    
\end{equation}

Here, \(\mathcal{P}\) denotes the set of positive gene pairs (genes in the same pathway), and \(\mathcal{N}(i)\) includes all gene pairs involving gene \(i\) in the batch, including negatives. The numerator in the logarithm encourages high similarity for positive pairs, while the denominator includes all other pairs, ensuring the model contrasts between intra-pathway and inter-pathway relationships. This loss function ensures that embeddings of genes within the same pathway are pulled closer together, capturing their shared biological context, while embeddings of genes from different pathways are pushed apart. By leveraging pathway information, this approach enhances the biological interpretability and functional organization of gene embeddings. The final pretraning loss is the weighted sum of the three loss:

\begin{equation}
\mathcal{L} = \mathcal{L}_{\text{lang}} + \gamma \mathcal{L}_{\text{pathway}}    
\end{equation}

\section{Experiments}

\textbf{Pretraining Dataset Construction}

Our pretrianing dataset was constructed using single-cell RNA sequencing (scRNA-seq) data sourced from the CELLXGENE database, including raw count matrices and their corresponding metadata. The original dataset comprised 50,689,395 cells. To ensure data quality, duplicates were removed by retaining only entries where the feature is\_primary\_data was set to True. This step eliminated approximately 41\% of the samples, resulting in 30,139,066 unique cells. Subsequently, the data was filtered to include 22,053 human protein-coding or miRNA genes. Cells expressing fewer than 200 genes were excluded to remove low-quality samples. The total gene expression counts for each cell were normalized to a fixed target value, and a logarithmic transformation ($log1p$) was applied to scale the data and reduce skewness. Following preprocessing, the dataset used for pretraining was finalized with 29,849,897 cells. For further refinement, a normalization factor was computed for each gene by calculating the non-zero median expression value of that gene across all cells. Genes were then ranked based on their normalized expression levels within each cell, and the top 2,048 or 4,096 gene indices were selected as input features for the pretraining stage. The $\gamma$ value is set to 0.1 in our experiments based on validation results from sample datasets. The code and datasets are available at an anonymous address: https://anonymous.4open.science/r/GeneMamba-747D.

\textbf{Pretraining Configuration}
We configured GeneMamba with 24 Bi-Mamba layers, an inner dimension of 512, and a vocabulary size of 25,426, resulting in 65.74 million learnable parameters. The model was trained for five epochs, distributed across four NVIDIA A100-SXM4-80GB GPUs, requiring approximately three weeks to complete. Additionally, we implemented variations of GeneMamba and compared their performance.

\textbf{Downstream Task Datasets}
For the downstream tasks in the GeneMamba paper, we utilized several datasets tailored to specific evaluations. For cell type annotation, we employed hPancreas, MS, Myeloid, and Myeloid\_b, with Myeloid\_b derived from Myeloid by excluding extremely small cell types to ensure robust performance assessment. Multi-batch analysis was conducted using the PBMC12k, COVID-19, and Perirhinal Cortex datasets to evaluate the model's effectiveness across diverse batch settings. For gene correlation analysis, the Immune and BMMC datasets were utilized to explore gene-gene relationships and validate the model’s ability to capture meaningful biological insights. More details can be found in Appendix~\ref{section:appendix_downstream}.

\textbf{Baselines}
We evaluate our model against established baselines, including GeneFormer, scGPT, scFoundation, and scBert, which are transformer-based models designed for single-cell data analysis. Additionally, we compare with Harmony, a widely used computational biology toolkit for clustering and cell-type classification. These methods provide a comprehensive benchmark, spanning deep learning and traditional computational biology approaches. More details about the baseline models are introduced in Appendix~\ref{section:baselines}

\subsection{Multi-batch Integration Task}

\begin{figure}[ht]
\begin{center}
\centerline{\includegraphics[width=\columnwidth]{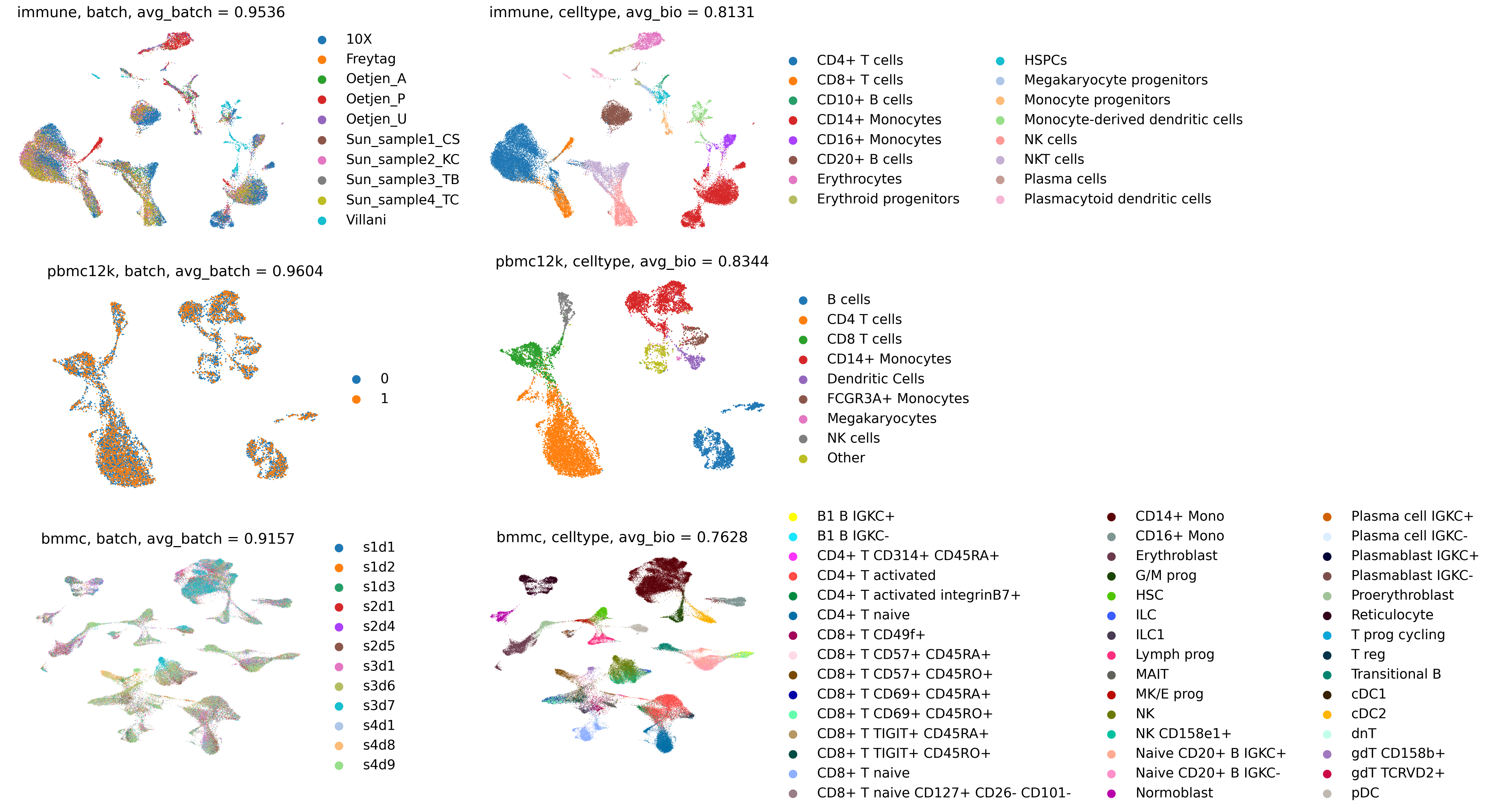}}
\caption{Results of multi-batch integration. Benchmark of the fine-tuned GeneMamba on the PBMC 12k dataset for the multi-batch integration task. The UMAP plot of learned cell embeddings is colored by cell types.}
\label{fig:multi_batch}
\end{center}
\end{figure}
In this experiment, we evaluated the multi-batch integration performance of the GeneMamba model. Multi-batch integration refers to aligning and harmonizing single-cell datasets from multiple experimental batches to minimize batch effects while preserving biologically meaningful patterns. We fine-tuned the pretrained GeneMamba model by adding a simple classification head (MLP) on the PBMC12k, COVID-19, and perirhinal cortex datasets. Using the learned embeddings from these foundation models, we performed multi-batch integration experiments.

The Avg-batch metric assesses the model's ability to correct batch effects and integrate data across multiple batches, while the Avg-bio metric evaluates the preservation of biological variation and meaningful clustering of cell types after integration. As shown in Table~\ref{tab:multi_batch_integration} and Figure~\ref{fig:multi_batch}, batch correction and biological preservation often present a trade-off. Harmony, as a specialized method for batch effect correction, excels in eliminating batch effects but tends to ignore biological differences. In contrast, GeneMamba demonstrates superior performance by effectively mitigating batch effects while preserving biological information. This highlights its robust capability in multi-batch integration and its biological relevance for single-cell data analysis. For further details, please refer to Figures~\ref{fig:appendix_multi_batch_covid19} and~\ref{fig:appendix_multi_batch_percor}.

\begin{table*}[h]
\centering
\caption{Benchmark results of the models on multi-batch experiments with BATCH and Cell metrics}
\resizebox{\textwidth}{!}{%
\begin{tabular}{llcccccc}
\toprule
\textbf{Metric} & \textbf{Dataset} & \textbf{Harmony} & \textbf{GeneFormer} & \textbf{scGPT} & \textbf{scFoundation} & \textbf{GeneMamba} \\
\midrule
\multirow{5}{*}{\textbf{Avg\_batch}} 
& Immune            & 0.9514         & 0.8153              & 0.9194         & 0.8904               & \textbf{0.9536}             \\
& PBMC12k           & 0.9341         & 0.9545              & \textbf{0.9755}         & 0.9628               & 0.9604             \\
& BMMC              & 0.8999         & 0.7720              & 0.8431         & 0.7598               & \textbf{0.9157}             \\
& Perirhinal Cortex & 0.9442        & 0.9127              & 0.9600         & 0.9560               & \textbf{0.9673}             \\
& Covid-19          & \textbf{0.8781}         & 0.8240              & 0.8625         & 0.8346               & 0.8742             \\
\midrule
\multirow{5}{*}{\textbf{Avg\_bio}} 
& Immune            & 0.6945         & 0.6983              & 0.7879         & 0.7337               & \textbf{0.8131}             \\
& PBMC12k           & 0.7990         & 0.7891              & \textbf{0.9018}         & 0.8662               & 0.8344             \\
& BMMC              & 0.6316         & 0.6324              & 0.6576         & 0.5250               & \textbf{0.7628}             \\
& Perirhinal Cortex & 0.8595       & 0.8547              & 0.9552         & \textbf{0.9606}               & 0.9062             \\
& Covid-19          & 0.4468         & 0.5567              & \textbf{0.6476}         & 0.5468               & 0.5537             \\
\bottomrule
\end{tabular}
}
\label{tab:multi_batch_integration}
\end{table*}

\subsection{Cell Type Annotation}

In this experiment, we evaluate the classification ability of the GeneMamba model on cell type annotation tasks using four benchmark datasets: hPancreas, MS, Myeloid, and Myeloid\_b. The Myeloid\_b dataset is a modified version of the Myeloid dataset, created by excluding extremely small cell types to ensure a more balanced distribution of cell populations. These datasets encompass diverse biological contexts, allowing for a comprehensive evaluation of the model’s capabilities. Detailed statistics and characteristics of these datasets are provided in Appendix \ref{section:appendix_downstream} for reference.

The annotation experiment employs the GeneMamba model fine-tuned with a simple classification head (MLP) in a supervised learning setup, directly testing its ability to learn and predict cell types based on input data. As summarized in Table~\ref{tab:cell_type_annotation_comparison_results}, the GeneMamba model demonstrates competitive performance across datasets. It achieves the highest accuracy (0.9713) and Macro-F1 (0.7710) scores for the hPancreas dataset, showing its robustness in capturing complex cell-type variations. Similarly, in the Myeloid dataset, GeneMamba outperforms others with a Macro-F1 score of 0.3650, showcasing its edge in identifying nuanced differences between cell types. While it lags in the MS dataset, its performance on the balanced Myeloid\_b dataset (Acc: 0.9603, Macro-F1: 0.9235) remains strong, underscoring its capability in handling challenging tasks.

These results highlight the effectiveness of GeneMamba’s architecture and training strategy in capturing meaningful biological patterns for precise cell type classification. The embeddings produced by the model capture intrinsic differences across cell types, simplifying downstream tasks and enabling the use of a simple classifier to achieve high accuracy. Furthermore, the confusion matrix in Appendix Figure~\ref{fig:cm_four_datasets} reveals clear patterns of prediction accuracy and misclassification, reflecting GeneMamba’s robustness and precision in distinguishing between diverse cell types.

\begin{table}[h]
\centering
\caption{Benchmark annotation performance across datasets}
\renewcommand{\arraystretch}{1.2}
\setlength{\tabcolsep}{5pt}
\begin{minipage}[t]{0.48\textwidth}
\centering
\resizebox{0.95\textwidth}{!}{%
\begin{tabular}{llcc}
\toprule
\textbf{Datasets} & \textbf{Models} & \textbf{Acc} & \textbf{Macro-F1} \\
\midrule
\multirow{4}{*}{hPancreas}
  & GeneFormer     & 0.9665 & 0.7450 \\
  & scGPT          & 0.9710 & 0.7632 \\
  & scFoundation   & 0.9602 & 0.7101 \\
  & GeneMamba      & \textbf{0.9713} & \textbf{0.7710} \\
\midrule
\multirow{4}{*}{MS}
  & GeneFormer     & 0.7650 & 0.6220 \\
  & scGPT          & \textbf{0.8471} & 0.6630 \\
  & scFoundation   & 0.7763 & \textbf{0.6812} \\
  & GeneMamba      & 0.6825 & 0.5342 \\
\bottomrule
\end{tabular}
}
\end{minipage}
\hfill
\begin{minipage}[t]{0.48\textwidth}
\centering
\resizebox{0.95\textwidth}{!}{%
\begin{tabular}{llcc}
\toprule
\textbf{Datasets} & \textbf{Models} & \textbf{Acc} & \textbf{Macro-F1} \\
\midrule
\multirow{4}{*}{Myeloid}
  & GeneFormer     & 0.6445 & 0.3600 \\
  & scGPT          & 0.6341 & 0.3562 \\
  & scFoundation   & 0.6446 & \textbf{0.3646} \\
  & GeneMamba      & \textbf{0.6607} & 0.3650 \\
\midrule
\multirow{4}{*}{Myeloid\_b}
  & GeneFormer     & 0.9540 & 0.9380 \\
  & scGPT          & 0.9421 & 0.9434 \\
  & scFoundation   & 0.9574 & \textbf{0.9569} \\
  & GeneMamba      & \textbf{0.9603} & 0.9235 \\
\bottomrule
\end{tabular}
}
\end{minipage}

\label{tab:cell_type_annotation_comparison_results}
\end{table}



\subsection{Gene Rank Reconstruction}

\begin{wrapfigure}{r}{0.5\textwidth}
\centering
\includegraphics[width=1.0\linewidth]{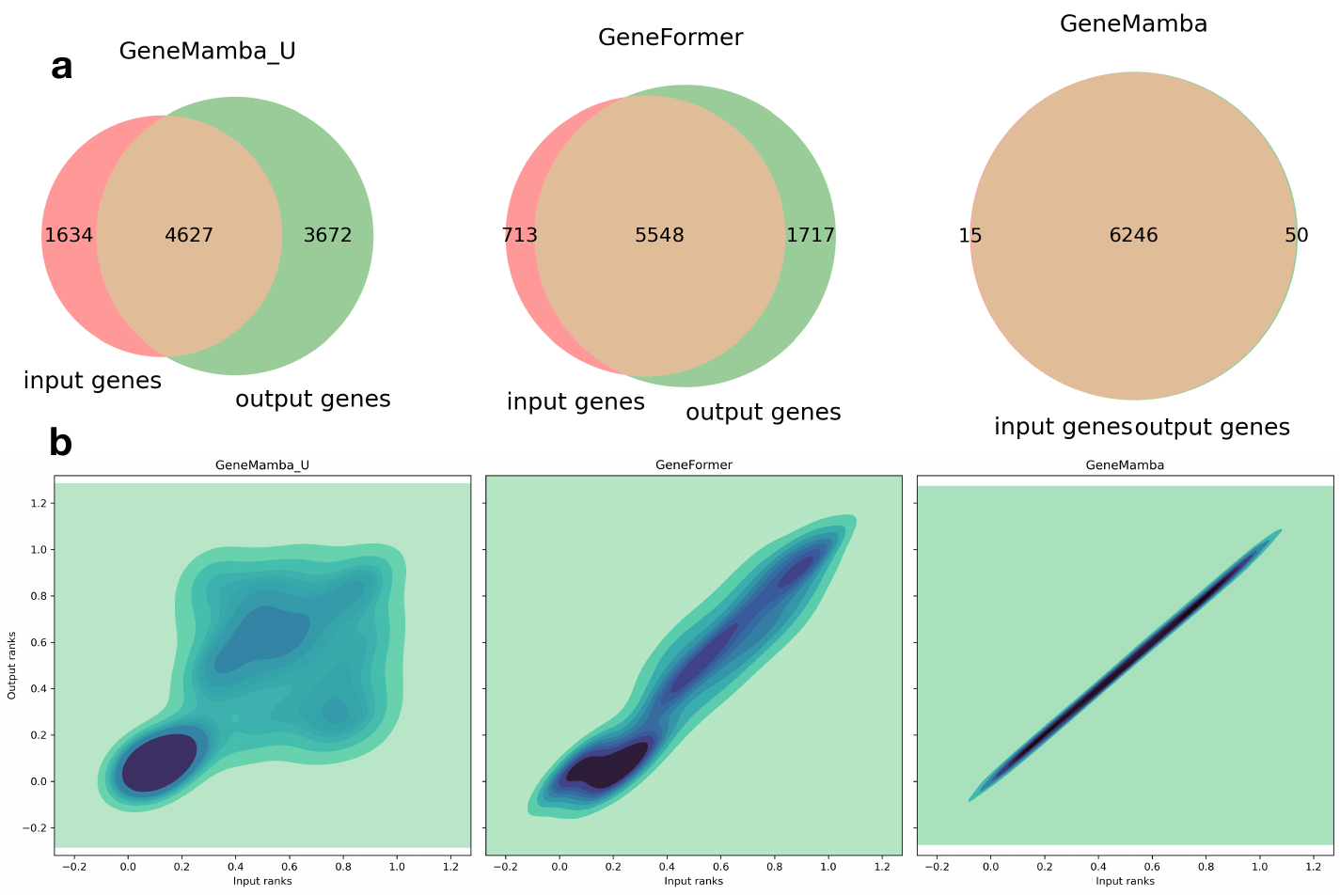}
\caption{Gene rank reconstruct results on PBMC12k dataset. (a) Venn diagrams showing overlapping between input and output tokens in the pancreas dataset by three models: GeneMamba\_U (unidirectional Mamba module as backbone), GeneFormer, GeneMamba (BiMamba module as backbone). (b) Density plots showing input and output ranking in the pancreas dataset by three models: GeneMamba\_U, GeneFormer, GeneMamba.}
\label{fig:reconstruct_result_pbmc12k}
\end{wrapfigure}

The GeneMamba model exhibits strong capability in reconstructing ranked gene orders, a crucial requirement for single-cell analysis tasks where gene expression patterns reflect cellular states and transitions. This challenge has been highlighted in prior studies \cite{kedzierska2023assessing, boiarsky2023deep}, and here we demonstrate the effectiveness of GeneMamba in addressing it. The analysis begins with the random selection of sample cells, from which input gene tokens are extracted. GeneMamba then generates output tokens, which are ranked according to their predicted likelihoods. To maintain sequence consistency, missing tokens are imputed with zeros. This strategy preserves the structural integrity of the data and enables a direct comparison between the input and output rankings.


The consistency of GeneMamba’s predictions is validated through a Venn diagram Figure~\ref{fig:reconstruct_result_pbmc12k}a, which illustrates a high degree of overlap between input and output tokens for the pancreas dataset. This overlap highlights the model's ability to retain critical features of the input data, an essential characteristic for single-cell studies where the integrity of gene expression ranks directly influences downstream analyses. A density plot Figure~\ref{fig:reconstruct_result_pbmc12k}b provides further evidence of rank fidelity. It demonstrates that lower-ranked input genes consistently yield lower-ranked outputs, indicating a linear relationship between input and output ranks. This observation underscores the model's effectiveness in capturing patterns within the data and maintaining rank consistency.

\begin{table}
\centering
\caption{Gene rank reconstruction performance on the PBMC12k dataset}
\begin{tabularx}{\columnwidth}{l>{\centering\arraybackslash}X>{\centering\arraybackslash}X>{\centering\arraybackslash}X}
\hline
         Model &   L-Dist &   BLEU &   Spearman \\
\hline
        GeneMamba\_U & 430 & 0.532 & 0.469 \\
        GeneFormer & 23 & 0.968 & 0.703 \\
        GeneMamba & \textbf{6} & \textbf{0.987} & \textbf{0.711} \\
\hline
\end{tabularx}
\label{tab:reconstruct_result_pbmc12k}
\end{table}

Comparative analysis (Figure~\ref{fig:reconstruct_result_pbmc12k} from left to right) reveals the advantages of GeneMamba over other models: (1) The inclusion of the BiMamba module significantly enhances performance compared to the unidirectional module. This result demonstrates the necessity of bidirectional context extraction for effectively reconstructing ranked gene orders. (2) The GeneFormer model excels at predicting higher-ranked genes but struggles with lower-ranked genes, often treating them as noise due to their lower frequency. In contrast, GeneMamba captures both higher- and lower-ranked genes effectively, showcasing its robustness and sensitivity to varying gene expression levels. The metric results are summarized in Table~\ref{tab:reconstruct_result_pbmc12k}. Additional results are provided in the Appendix Table~\ref{tab:appendix_gene_rank_reconstruct_models_comparison}.

\vspace{-0.8em}
\section{Conclusion}
\vspace{-0.8em}
We present GeneMamba, a scalable model for single-cell analysis that treats cells as sentences and genes as tokens. Using the BiMamba block and biologically informed losses, GeneMamba captures gene relationships and cell diversity effectively. It performs well across key tasks and offers a strong foundation for future single-cell studies.

\textbf{Broader Impacts and Limitations.} GeneMamba can help researchers analyze large-scale single-cell data, offering insights into gene function and cell behavior. This has potential benefits in areas like disease research and drug development. However, the model may struggle with rare cell types or subtle signals from low-expression genes. It also requires significant computing resources for pretraining, which could limit broader use. Future work could focus on improving efficiency and expanding to more flexible architectures.

\bibliographystyle{plain}  
\bibliography{references}

@article{norman2019exploring,
  title={Exploring genetic interaction manifolds constructed from rich single-cell phenotypes},
  author={Norman, Thomas M and Horlbeck, Max A and Replogle, Joseph M and Ge, Alex Y and Xu, Albert and Jost, Marco and Gilbert, Luke A and Weissman, Jonathan S},
  journal={Science},
  volume={365},
  number={6455},
  pages={786--793},
  year={2019},
  publisher={American Association for the Advancement of Science}
}

@article{kedzierska2023assessing,
  title={Assessing the limits of zero-shot foundation models in single-cell biology},
  author={Kedzierska, Kasia Z and Crawford, Lorin and Amini, Ava P and Lu, Alex X},
  journal={bioRxiv},
  pages={2023--10},
  year={2023},
  publisher={Cold Spring Harbor Laboratory}
}

@article{scTCA,
  title={scTCA: a hybrid Transformer-CNN architecture for imputation and denoising of scDNA-seq data},
  author={Yu, Zhenhua and Liu, Furui and Li, Yang},
  journal={Briefings in Bioinformatics},
  volume={25},
  number={6},
  pages={bbae577},
  year={2024},
  publisher={Oxford University Press}
}

@article{scHyena,
  title={scHyena: Foundation Model for Full-Length Single-Cell RNA-Seq Analysis in Brain},
  author={Gyutaek Oh and Baekgyu Choi and Inkyung Jung and Jong Chul Ye},
  journal={arXiv preprint arXiv:2310.02713},
  year={2023}
}

@article{scTransSort,
  title={scTransSort: Transformers for intelligent annotation of cell types by gene embeddings},
  author={Jiao, Linfang and Wang, Gan and Dai, Huanhuan and Li, Xue and Wang, Shuang and Song, Tao},
  journal={Biomolecules},
  volume={13},
  number={4},
  pages={611},
  year={2023},
  publisher={MDPI}
}

@article{scGraphformer,
  title={scGraphformer: unveiling cellular heterogeneity and interactions in scRNA-seq data using a scalable graph transformer network},
  author={Fan, Xingyu and Liu, Jiacheng and Yang, Yaodong and Gu, Chunbin and Han, Yuqiang and Wu, Bian and Jiang, Yirong and Chen, Guangyong and Heng, Pheng-Ann},
  journal={Communications Biology},
  volume={7},
  number={1},
  pages={1463},
  year={2024},
  publisher={Nature Publishing Group UK London}
}

@article{scGFT,
  title={Single-cell RNA-seq data augmentation using generative Fourier transformer},
  author={Nouri, Nima},
  journal={Communications Biology},
  volume={8},
  number={1},
  pages={113},
  year={2025},
  publisher={Nature Publishing Group UK London}
}

@article{WhiteBoxDiffusionTransformer,
  title={White-Box Diffusion Transformer for single-cell RNA-seq generation},
  author={Cui, Zhuorui and Dong, Shengze and Liu, Ding},
  journal={arXiv preprint arXiv:2411.06785},
  year={2024}
}

@article{STGRNS,
  title={STGRNS: an interpretable transformer-based method for inferring gene regulatory networks from single-cell transcriptomic data},
  author={Xu, Jing and Zhang, Aidi and Liu, Fang and Zhang, Xiujun},
  journal={Bioinformatics},
  volume={39},
  number={4},
  pages={btad165},
  year={2023},
  publisher={Oxford University Press}
}

@article{scRDiT,
  title={scRDiT: Generating single-cell RNA-seq data by diffusion transformers and accelerating sampling},
  author={Shengze Dong and Zhuorui Cui and Ding Liu and Jinzhi Lei},
  journal={arXiv preprint arXiv:2404.06153},
  year={2024}
}

@article{TransformerST,
  title={Innovative super-resolution in spatial transcriptomics: a transformer model exploiting histology images and spatial gene expression},
  author={Zhao, Chongyue and Xu, Zhongli and Wang, Xinjun and Tao, Shiyue and MacDonald, William A and He, Kun and Poholek, Amanda C and Chen, Kong and Huang, Heng and Chen, Wei},
  journal={Briefings in Bioinformatics},
  volume={25},
  number={2},
  pages={bbae052},
  year={2024},
  publisher={Oxford University Press}
}

@article{oord2018representation,
  title={Representation learning with contrastive predictive coding},
  author={Oord, Aaron van den and Li, Yazhe and Vinyals, Oriol},
  journal={arXiv preprint arXiv:1807.03748},
  year={2018}
}

@article{garcia2015pathway,
  title={Pathway analysis: state of the art},
  author={Garc{\'\i}a-Campos, Miguel A and Espinal-Enr{\'\i}quez, Jes{\'u}s and Hern{\'a}ndez-Lemus, Enrique},
  journal={Frontiers in physiology},
  volume={6},
  pages={383},
  year={2015},
  publisher={Frontiers Media SA}
}

@inproceedings{zien2000analysis,
  title={Analysis of gene expression data with pathway scores.},
  author={Zien, Alexander and K{\"u}ffner, Robert and Zimmer, Ralf and Lengauer, Thomas},
  booktitle={Ismb},
  volume={8},
  pages={407--417},
  year={2000}
}

@article{gundogdu2022integrating,
  title={Integrating pathway knowledge with deep neural networks to reduce the dimensionality in single-cell RNA-seq data},
  author={Gundogdu, Pelin and Loucera, Carlos and Alamo-Alvarez, Inmaculada and Dopazo, Joaquin and Nepomuceno, Isabel},
  journal={BioData Mining},
  volume={15},
  pages={1--21},
  year={2022},
  publisher={Springer}
}

@article{sharma2023regenne,
  title={ReGeNNe: genetic pathway-based deep neural network using canonical correlation regularizer for disease prediction},
  author={Sharma, Divya and Xu, Wei},
  journal={Bioinformatics},
  volume={39},
  number={11},
  pages={btad679},
  year={2023},
  publisher={Oxford University Press}
}

@article{li2024single,
  title={Single-cell RNA sequencing in cancer research: New insights and applications},
  author={Li, Jing and Wang, Yu},
  journal={Cancer Research},
  volume={84},
  number={10},
  pages={1234--1245},
  year={2024},
  publisher={American Association for Cancer Research}
}

@article{deconinck2021recent,
  title={Recent advances in trajectory inference from single-cell omics data},
  author={Deconinck, Louise and Cannoodt, Robrecht and Saelens, Wouter and Deplancke, Bart and Saeys, Yvan},
  journal={Current Opinion in Systems Biology},
  volume={27},
  pages={100344},
  year={2021},
  publisher={Elsevier}
}

@article{petegrosso2020machine,
  title={Machine learning and statistical methods for clustering single-cell RNA-sequencing data},
  author={Petegrosso, Raphael and Li, Zhuliu and Kuang, Rui},
  journal={Briefings in bioinformatics},
  volume={21},
  number={4},
  pages={1209--1223},
  year={2020},
  publisher={Oxford University Press}
}

@article{ryu2023integration,
  title={Integration of single-cell RNA-seq datasets: a review of computational methods},
  author={Ryu, Yeonjae and Han, Geun Hee and Jung, Eunsoo and Hwang, Daehee},
  journal={Molecules and cells},
  volume={46},
  number={2},
  pages={106--119},
  year={2023},
  publisher={Elsevier}
}

@article{white2024comprehensive,
  title={A comprehensive review of computational methods for scRNA-seq},
  author={White, Daniel and Harris, Olivia},
  journal={Briefings in Bioinformatics},
  volume={25},
  number={6},
  pages={789--799},
  year={2024},
  publisher={Oxford University Press}
}

@InProceedings{mamba2,
  title = 	 {Transformers are {SSM}s: Generalized Models and Efficient Algorithms Through Structured State Space Duality},
  author =       {Dao, Tri and Gu, Albert},
  booktitle = 	 {Proceedings of the 41st International Conference on Machine Learning},
  pages = 	 {10041--10071},
  year = 	 {2024},
  editor = 	 {Salakhutdinov, Ruslan and Kolter, Zico and Heller, Katherine and Weller, Adrian and Oliver, Nuria and Scarlett, Jonathan and Berkenkamp, Felix},
  volume = 	 {235},
  series = 	 {Proceedings of Machine Learning Research},
  month = 	 {21--27 Jul},
  publisher =    {PMLR}
}

@article{korsunsky2019fast,
  title={Fast, sensitive and accurate integration of single-cell data with Harmony},
  author={Korsunsky, Ilya and Millard, Nghia and Fan, Jean and Slowikowski, Kamil and Zhang, Fan and Wei, Kevin and Baglaenko, Yuriy and Brenner, Michael and Loh, Po-ru and Raychaudhuri, Soumya},
  journal={Nature methods},
  volume={16},
  number={12},
  pages={1289--1296},
  year={2019},
  publisher={Nature Publishing Group US New York}
}

@article{du2019gene2vec,
  title={Gene2vec: distributed representation of genes based on co-expression},
  author={Du, Jingcheng and Jia, Peilin and Dai, Yulin and Tao, Cui and Zhao, Zhongming and Zhi, Degui},
  journal={BMC genomics},
  volume={20},
  pages={7--15},
  year={2019},
  publisher={Springer}
}

@article{theodoris2023transfer,
  title={Transfer learning enables predictions in network biology},
  author={Theodoris, Christina V and Xiao, Ling and Chopra, Anant and Chaffin, Mark D and Al Sayed, Zeina R and Hill, Matthew C and Mantineo, Helene and Brydon, Elizabeth M and Zeng, Zexian and Liu, X Shirley and others},
  journal={Nature},
  volume={618},
  number={7965},
  pages={616--624},
  year={2023},
  publisher={Nature Publishing Group UK London}
}

@article{yang2022scbert,
  title={scBERT as a large-scale pretrained deep language model for cell type annotation of single-cell RNA-seq data},
  author={Yang, Fan and Wang, Wenchuan and Wang, Fang and Fang, Yuan and Tang, Duyu and Huang, Junzhou and Lu, Hui and Yao, Jianhua},
  journal={Nature Machine Intelligence},
  volume={4},
  number={10},
  pages={852--866},
  year={2022},
  publisher={Nature Publishing Group UK London}
}

@inproceedings{bian2024scmulan,
  title={scMulan: a multitask generative pre-trained language model for single-cell analysis},
  author={Bian, Haiyang and Chen, Yixin and Dong, Xiaomin and Li, Chen and Hao, Minsheng and Chen, Sijie and Hu, Jinyi and Sun, Maosong and Wei, Lei and Zhang, Xuegong},
  booktitle={International Conference on Research in Computational Molecular Biology},
  pages={479--482},
  year={2024},
  organization={Springer}
}

@article{chen2023transformer,
  title={Transformer for one stop interpretable cell type annotation},
  author={Chen, Jiawei and Xu, Hao and Tao, Wanyu and Chen, Zhaoxiong and Zhao, Yuxuan and Han, Jing-Dong J},
  journal={Nature Communications},
  volume={14},
  number={1},
  pages={223},
  year={2023},
  publisher={Nature Publishing Group UK London}
}

@article{szalata2024transformers,
  title={Transformers in single-cell omics: a review and new perspectives},
  author={Sza{\l}ata, Artur and Hrovatin, Karin and Becker, S{\"o}ren and Tejada-Lapuerta, Alejandro and Cui, Haotian and Wang, Bo and Theis, Fabian J},
  journal={Nature methods},
  volume={21},
  number={8},
  pages={1430--1443},
  year={2024},
  publisher={Nature Publishing Group US New York}
}

@article{shen2022universal,
  title={A universal approach for integrating super large-scale single-cell transcriptomes by exploring gene rankings},
  author={Shen, Hongru and Shen, Xilin and Feng, Mengyao and Wu, Dan and Zhang, Chao and Yang, Yichen and Yang, Meng and Hu, Jiani and Liu, Jilei and Wang, Wei and others},
  journal={Briefings in Bioinformatics},
  volume={23},
  number={2},
  pages={bbab573},
  year={2022},
  publisher={Oxford University Press}
}

@article{shen2023generative,
  title={Generative pretraining from large-scale transcriptomes for single-cell deciphering},
  author={Shen, Hongru and Liu, Jilei and Hu, Jiani and Shen, Xilin and Zhang, Chao and Wu, Dan and Feng, Mengyao and Yang, Meng and Li, Yang and Yang, Yichen and others},
  journal={Iscience},
  volume={26},
  number={5},
  year={2023},
  publisher={Elsevier}
}

@inproceedings{xiong2023scclip,
  title={scCLIP: Multi-modal Single-cell Contrastive Learning Integration Pre-training},
  author={Xiong, Lei and Chen, Tianlong and Kellis, Manolis},
  year={2023},
  booktitle={NeurIPS 2023 AI for Science Workshop}
}

@article{zhao2023large,
  title={Large-scale cell representation learning via divide-and-conquer contrastive learning},
  author={Zhao, Suyuan and Zhang, Jiahuan and Nie, Zaiqing},
  journal={arXiv preprint arXiv:2306.04371},
  year={2023}
}

@article{wen2023cellplm,
  title={CellPLM: pre-training of cell language model beyond single cells},
  author={Wen, Hongzhi and Tang, Wenzhuo and Dai, Xinnan and Ding, Jiayuan and Jin, Wei and Xie, Yuying and Tang, Jiliang},
  journal={bioRxiv},
  pages={2023--10},
  year={2023},
  publisher={Cold Spring Harbor Laboratory}
}

@article{wen2023single,
  title={Single cells are spatial tokens: Transformers for spatial transcriptomic data imputation},
  author={Wen, Hongzhi and Tang, Wenzhuo and Jin, Wei and Ding, Jiayuan and Liu, Renming and Dai, Xinnan and Shi, Feng and Shang, Lulu and Liu, Hui and Xie, Yuying},
  journal={arXiv preprint arXiv:2302.03038},
  year={2023}
}

@article{zhao2024langcell,
  title={Langcell: Language-cell pre-training for cell identity understanding},
  author={Zhao, Suyuan and Zhang, Jiahuan and Wu, Yushuai and Luo, Yizhen and Nie, Zaiqing},
  journal={arXiv preprint arXiv:2405.06708},
  year={2024}
}

@article{cui2024scgpt,
  title={scGPT: toward building a foundation model for single-cell multi-omics using generative AI},
  author={Cui, Haotian and Wang, Chloe and Maan, Hassaan and Pang, Kuan and Luo, Fengning and Duan, Nan and Wang, Bo},
  journal={Nature Methods},
  pages={1--11},
  year={2024},
  publisher={Nature Publishing Group US New York}
}

@article{gong2024xtrimogene,
  title={xTrimoGene: an efficient and scalable representation learner for single-cell RNA-seq data},
  author={Gong, Jing and Hao, Minsheng and Cheng, Xingyi and Zeng, Xin and Liu, Chiming and Ma, Jianzhu and Zhang, Xuegong and Wang, Taifeng and Song, Le},
  journal={Advances in Neural Information Processing Systems},
  volume={36},
  year={2024}
}

@article{hao2024large,
  title={Large-scale foundation model on single-cell transcriptomics},
  author={Hao, Minsheng and Gong, Jing and Zeng, Xin and Liu, Chiming and Guo, Yucheng and Cheng, Xingyi and Wang, Taifeng and Ma, Jianzhu and Zhang, Xuegong and Song, Le},
  journal={Nature Methods},
  pages={1--11},
  year={2024},
  publisher={Nature Publishing Group US New York}
}

@article{yang2024genecompass,
  title={GeneCompass: deciphering universal gene regulatory mechanisms with a knowledge-informed cross-species foundation model},
  author={Yang, Xiaodong and Liu, Guole and Feng, Guihai and Bu, Dechao and Wang, Pengfei and Jiang, Jie and Chen, Shubai and Yang, Qinmeng and Miao, Hefan and Zhang, Yiyang and others},
  journal={Cell Research},
  pages={1--16},
  year={2024},
  publisher={Springer Nature Singapore Singapore}
}

@article{boiarsky2023deep,
  title={A deep dive into single-cell RNA sequencing foundation models},
  author={Boiarsky, Rebecca and Singh, Nalini and Buendia, Alejandro and Getz, Gad and Sontag, David},
  journal={bioRxiv},
  pages={2023--10},
  year={2023},
  publisher={Cold Spring Harbor Laboratory}
}

@article{vaswani2017attention,
  title={Attention is all you need},
  author={Vaswani, Ashish and Shazeer, Noam and Parmar, Niki and Uszkoreit, Jakob and Jones, Llion and Gomez, Aidan N and Kaiser, {\L}ukasz and Polosukhin, Illia},
  journal={Advances in neural information processing systems},
  volume={30},
  year={2017}
}

@article{choromanski2020rethinking,
  title={Rethinking attention with performers},
  author={Choromanski, Krzysztof and Likhosherstov, Valerii and Dohan, David and Song, Xingyou and Gane, Andreea and Sarlos, Tamas and Hawkins, Peter and Davis, Jared and Mohiuddin, Afroz and Kaiser, Lukasz and others},
  journal={arXiv preprint arXiv:2009.14794},
  year={2020}
}

@article{dao2022flashattention,
  title={Flashattention: Fast and memory-efficient exact attention with io-awareness},
  author={Dao, Tri and Fu, Dan and Ermon, Stefano and Rudra, Atri and R{\'e}, Christopher},
  journal={Advances in Neural Information Processing Systems},
  volume={35},
  pages={16344--16359},
  year={2022}
}

@article{gu2023mamba,
  title={Mamba: Linear-time sequence modeling with selective state spaces},
  author={Gu, Albert and Dao, Tri},
  journal={arXiv preprint arXiv:2312.00752},
  year={2023}
}

@article{sun2024bidirectional,
  title={Bidirectional Mamba with Dual-Branch Feature Extraction for Hyperspectral Image Classification},
  author={Sun, Ming and Zhang, Jie and He, Xiaoou and Zhong, Yihe},
  journal={Sensors},
  volume={24},
  number={21},
  pages={6899},
  year={2024},
  publisher={MDPI}
}

@article{liu2024bidirectional,
  title={Bidirectional gated mamba for sequential recommendation},
  author={Liu, Ziwei and Liu, Qidong and Wang, Yejing and Wang, Wanyu and Jia, Pengyue and Wang, Maolin and Liu, Zitao and Chang, Yi and Zhao, Xiangyu},
  journal={arXiv preprint arXiv:2408.11451},
  year={2024}
}

@article{liang2024bi,
  title={Bi-Mamba4TS: Bidirectional Mamba for Time Series Forecasting},
  author={Liang, Aobo and Jiang, Xingguo and Sun, Yan and Lu, Chang},
  journal={arXiv preprint arXiv:2404.15772},
  year={2024}
}

\newpage
\appendix

{\Large \textbf{Appendix}}
\section{Dataset Construction and Setup}
\label{section:data_curation}

We constructed our pretraining dataset manually, which is sourced from the cellxgene database, acquiring single-cell RNA sequencing (scRNA-seq) data in raw count matrix format alongside the corresponding metadata. The metadata available on this platform were submitted by the original data contributors, and therefore, we consider it to be partially reliable. It is worth mentioning that, for the first time, we corrected the ``organ`` column by mapping it to the ``tissue`` column in the original collected data.

The raw data consisted of 50,689,395 cells downloaded from cellxgene. Following the guidelines provided by the platform, which notes that ``in some cases, data from the same cell exists in different datasets, therefore cells can be duplicated throughout CELLxGENE Discover and by extension the Census,'' we removed duplicates by filtering for entries where the feature \texttt{is\_primary\_data} was set to \texttt{True}. This step eliminated 20,550,329 samples, leaving a total of 30,139,066 unique cells.

Subsequently, we filtered the data to include only 22,053 human protein-coding or miRNA genes. To ensure data quality and reduce noise from poor-quality cells, we retained cells expressing at least 200 genes. The total counts of gene expression values in each cell were then normalized to a fixed target value to ensure comparability across cells. Following normalization, we applied a logarithmic transformation (\texttt{log1p}) to scale the data for the following analyses.

After preprocessing, the dataset for pretraining was finalized with 29,849,897 cells. We calculated a normalization factor for each gene by determining the non-zero median expression value of that gene across all cells. This normalization factor was consistently applied to both the pretraining corpus and all future datasets presented to the model. This approach deprioritized ubiquitously highly expressed housekeeping genes while highlighting genes with lower expression levels that contribute significantly to distinguishing cell states.

For the tokenization process, genes were ranked by their normalized expression values within each cell, and the resulting tokenized data were stored in the Huggingface Datasets format, retaining all non-zero gene entries. To provide a non-parametric representation of the transcriptome for each cell, we selected the top 2,048 or 4,096 gene indices as the input data to pretraining stage.

\section{Downstream Datasets, Tasks and Evaluation Metrics}
\subsection{Downstream Datasets}
\label{section:appendix_downstream}

The statistics of the datasets are shown in the Table ~\ref{tab:dataset_statistics}.

\begin{table}[h]
\centering
\caption{Statistics of downstream datasets, including the number of cells (\textit{n\_cells}), genes (\textit{n\_genes}), batches (\textit{n\_batches}), and cell types (\textit{n\_cell\_types}) for each dataset.}
\begin{tabular}{lcccc}
\hline
\textbf{Dataset} & \textbf{n\_cells} & \textbf{n\_genes} & \textbf{n\_batches} & \textbf{n\_cell\_types} \\
\hline
bmmc                & 90261           & 14087            & 12                & 45                \\
covid19             & 20000           & 1200             & 2                 & 39                \\
pbmc12k             & 11990           & 3346             & 2                 & 9                 \\
perirhinal cortex  & 17535           & 59357            & 2                 & 10                \\
immune              & 33506           & 12303            & 10                & 16                \\
myeloid             & 13178           & 3000             & 2                 & 21                \\
hpancreas           & 14818           & 3000             & 2                 & 14                \\
ms                  & 21312           & 3000             & 2                 & 18                \\
\hline
\end{tabular}
\label{tab:dataset_statistics}
\end{table}

\textbf{hPancreas}
This dataset comprises single-cell RNA sequencing (scRNA-seq) data from human pancreatic cells, including various cell types such as alpha, beta, delta, and acinar cells. It is utilized to study cellular heterogeneity and gene expression profiles within the pancreas.

\textbf{MS}
The multiple sclerosis (MS) dataset includes scRNA-seq data from peripheral blood mononuclear cells (PBMCs) of individuals with MS. It aids in understanding immune cell composition and gene expression changes associated with MS.

\textbf{Myeloid}
This dataset contains scRNA-seq data focusing on myeloid cells, such as monocytes and macrophages, from various tissues or conditions. It is essential for studying the role of myeloid cells in immune responses and diseases.

\textbf{PBMC12k}
The PBMC12k dataset consists of scRNA-seq data from 12,000 peripheral blood mononuclear cells obtained from a healthy donor. It serves as a reference for immune cell types and their gene expression profiles in the bloodstream.

\textbf{BMMC}
This dataset includes scRNA-seq data from bone marrow mononuclear cells, encompassing various hematopoietic cell types. It is utilized to study hematopoiesis and bone marrow microenvironments.

\textbf{COVID19}
The COVID19 dataset comprises scRNA-seq data from PBMCs of COVID-19 patients. It facilitates the investigation of immune responses and cellular changes during SARS-CoV-2 infection.

\textbf{Immune Human}
This dataset contains scRNA-seq data from human immune cells across different tissues or conditions. It aids in understanding the diversity and function of the human immune system.

\textbf{Perirhinal Cortex}
This dataset includes scRNA-seq data from cells of the perirhinal cortex, a region of the brain involved in memory and recognition. It is used to study neuronal and glial cell types and their gene expression profiles in this specific brain area.

\begin{figure}[h]
\begin{center}
\centerline{\includegraphics[width=0.6\columnwidth]{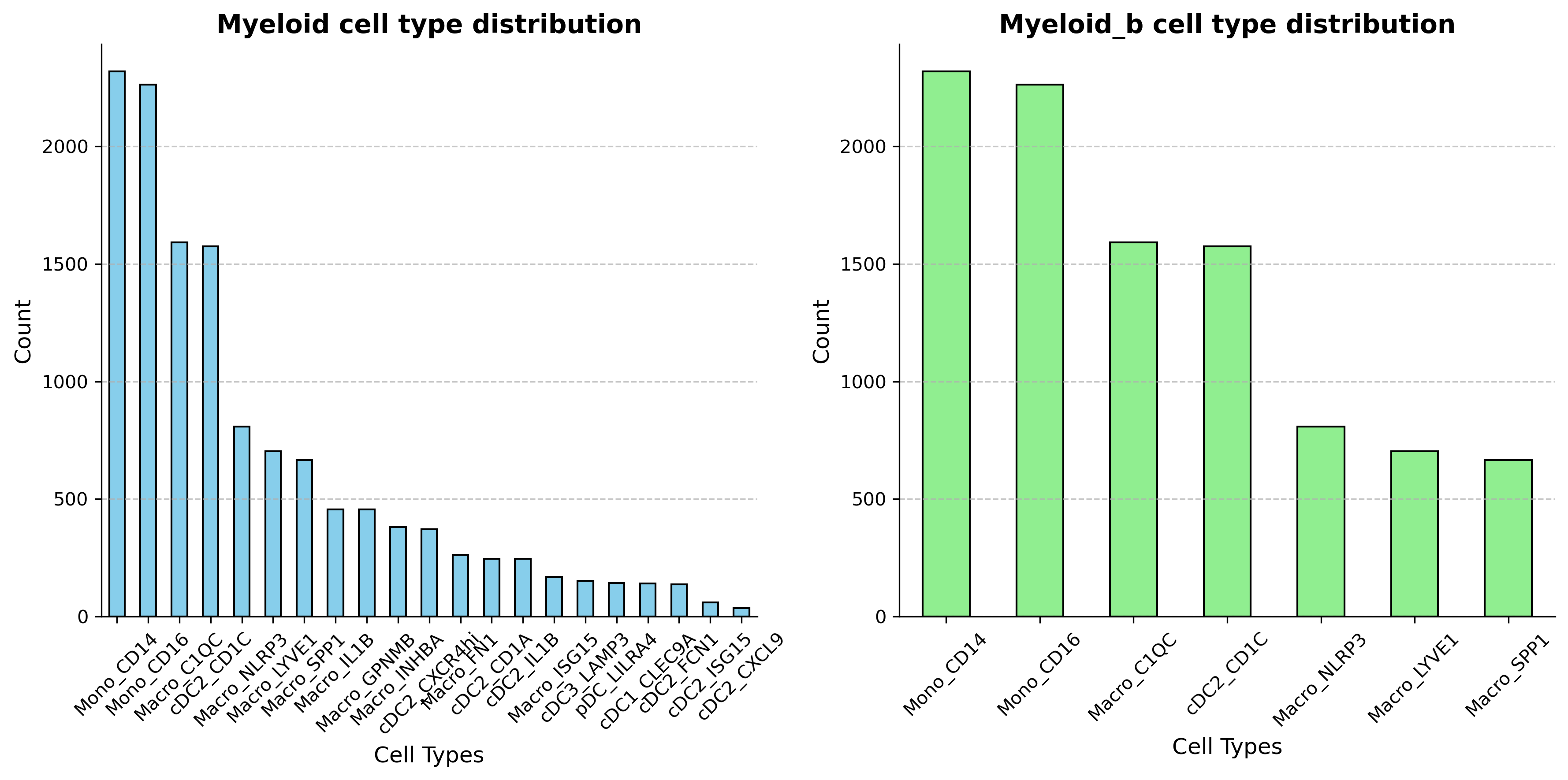}}
\caption{Cell type distribution in the original and modified Myeloid datasets}
\label{fig:myeloid_distribution}
\end{center}
\end{figure}

The original Myeloid dataset is highly imbalanced. To better evaluate the models' performance, we excluded rare cell types (proportion $<$ 0.05) to create the Myeloid\_b dataset. Figure~\ref{fig:myeloid_distribution} illustrates the cell type distribution in both the original and modified Myeloid datasets.

\subsection{Downstream Tasks}

We standardize the downstream datasets by converting the batch key to "batch" and the label key to "celltype." We then use a manual stratified split to divide the dataset into train/test with a 0.9/0.1 ratio to avoid missing rare classes in the test set. For example, in the COVID dataset, using `train\_test\_split()` directly would cause the 22nd class to be missing, as it contains only 3 samples, and a 0.3 test sample rate would result in zero samples, causing errors during AUC-ROC calculation. The split column is labeled "partition," where "train" is for training and "test" is for testing. 

\subsection*{Multi-batch integration Task}

Multi-batch integration is a task in single-cell data analysis that involves harmonizing data collected from multiple experimental batches to mitigate technical differences while preserving meaningful biological information. In this context, let \( X = \{X_1, X_2, \ldots, X_n\} \) denote datasets from \( n \) distinct batches, where \( X_i \in \mathbb{R}^{m_i \times p} \) represents the gene expression matrix of the \( i \)-th batch with \( m_i \) cells and \( p \) genes. The objective is to project these datasets into a unified latent space \( Z \in \mathbb{R}^{M \times d} \), where \( M = \sum_{i=1}^n m_i \) is the total number of cells across all batches, and \( d \) is the dimensionality of the latent space. In this shared space, cells with similar biological profiles are grouped together, regardless of their batch origin. This task is crucial for ensuring the comparability of cells from different batches and facilitating downstream analyses like clustering or cell type classification. Successful integration minimizes batch-specific artifacts and emphasizes biologically relevant signals, enabling a coherent and unbiased interpretation of single-cell data.

\subsection*{Cell Type Annotation}

Cell type annotation is a fundamental task in single-cell transcriptomics aimed at assigning biologically meaningful labels to individual cells based on their gene expression profiles. Formally, given a gene expression matrix \( X \in \mathbb{R}^{M \times p} \), where \( M \) is the total number of cells and \( p \) represents the number of genes, the goal is to map each cell \( x_i \in \mathbb{R}^p \) to a cell type label \( y_i \in \mathcal{Y} \), where \( \mathcal{Y} = \{y_1, y_2, \ldots, y_k\} \) is the set of \( k \) predefined cell types. This task leverages the distinct transcriptional signatures of cell types, often characterized by differential expression of marker genes, to provide insights into cellular identity and function. Successful annotation ensures that biologically relevant features are preserved, enabling accurate downstream analyses such as the study of cellular heterogeneity, lineage tracing, and disease-specific cell state identification.

\subsection*{Gene Rank Reconstruction}

\begin{figure}[h]
\begin{center}
\centerline{\includegraphics[width=0.6\columnwidth]{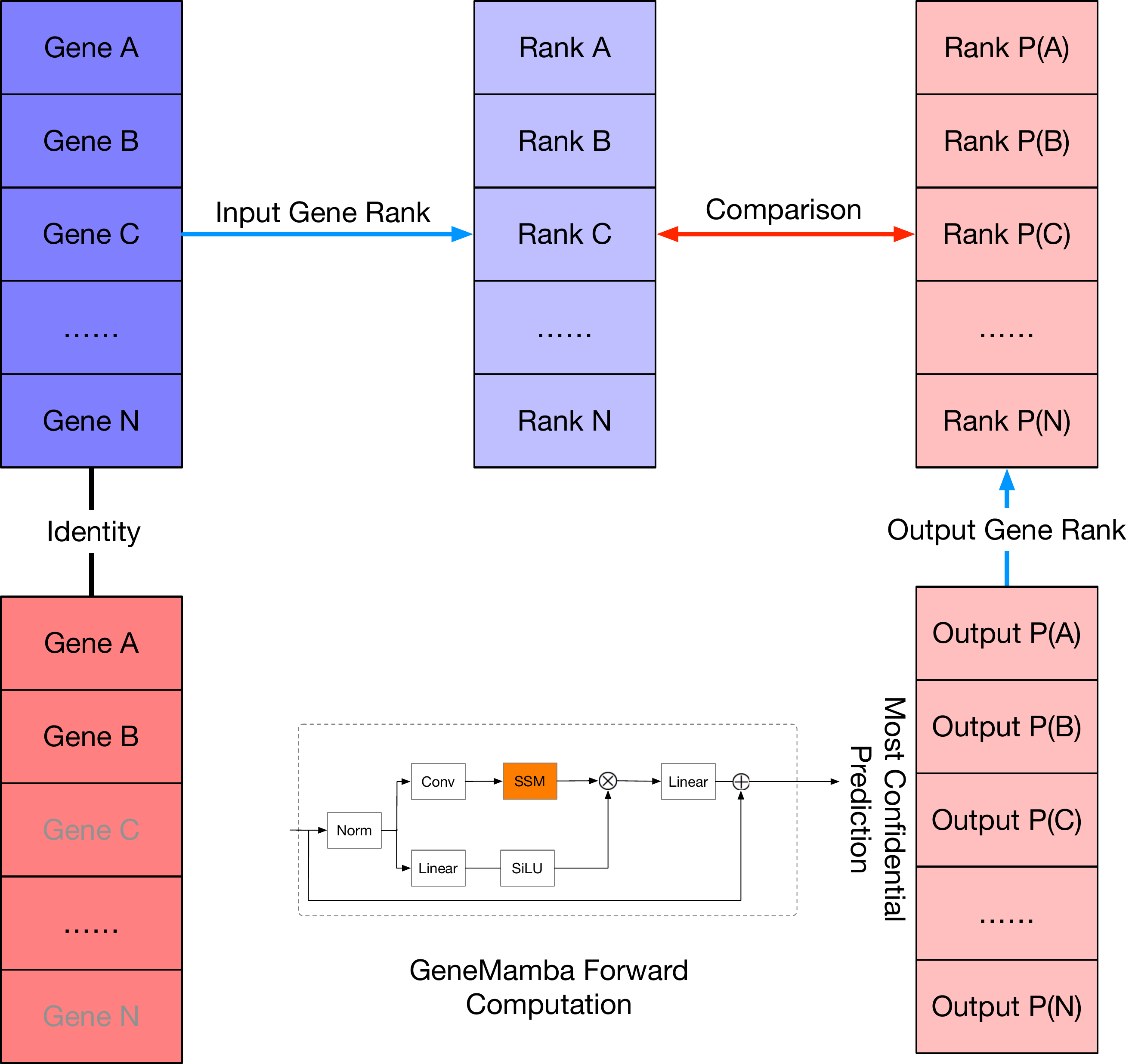}}
\caption{Gene Rank Reconstruct framework of GeneMamba}
\label{fig:reconstruct_framework}
\end{center}
\end{figure}

In the figure \ref{fig:reconstruct_framework}, the framework of doing the gene rank reconstruction experiment. The framework demonstrates the iterative nature of rank reconstruction, where the model processes one token at a time, maintaining sequential dependencies inherent in biological systems. This approach mirrors natural biological progressions, further emphasizing the model’s relevance to genomics.

\subsection*{Gene Distribution Alignment}

In this experiment, we evaluate the GeneMamba's performance to distinguish between positive pairs and negative pairs, the reference data is collected in the Gene2Vec paper~\cite{du2019gene2vec}. And we generate the embeddings using different models for positive gene pairs and negative gene pairs, then we measure the embeddings similarity of each pair. We evaluate how the distribution of positive pairs and negative pairs differ by calculating the eucilean distance, KL-Divergence, and JS-Divergence of the two distributions, the result is shown in Figure \ref{fig:gene_correlation_analysis}.

\subsection*{Gene Topology Preservation}
To evaluate gene-gene topology preservation, we conducted experiments using the pbmc12k dataset. A random selection of gene-gene pairs was made from the dataset, and their embeddings were obtained using three models: scGPT, GeneMamba, and Gene2Vec. The Euclidean distances between the gene pairs were computed, and an adjacency matrix was constructed to represent these relationships. The threshold for the distances was set as the mean value of the matrix elements, and the distances were binarized into 0 and 1 based on this threshold. To compare the topological similarity among models, the Jaccard distances between the adjacency matrices were calculated. Additionally, overlapping gene-gene pairs were identified to further understand shared topology. The experiment culminated in a visual representation of the gene-gene topology, highlighting the models' ability to preserve biological relationships in gene expression data.

\subsection{Evaluation Metrics}

\subsection*{AvgBIO}

To evaluate how well the integration preserves biological signals, we calculate the \textbf{AvgBIO} metric. This metric is the average of three biological conservation measures: the Adjusted Rand Index (\( ARI_\text{cell} \)), Normalized Mutual Information (\( NMI_\text{cell} \)), and the Average Silhouette Width based on cell types (\( ASW_\text{cell} \)):

\[
\text{AvgBIO} = \frac{ARI_\text{cell} + NMI_\text{cell} + ASW_\text{cell}}{3}.
\]

\textbf{1. Adjusted Rand Index (\( ARI_\text{cell} \)):  }
   We use \( ARI_\text{cell} \) to quantify the agreement between the true biological labels and the clusters predicted after integration. By adjusting for chance agreement, \( ARI_\text{cell} \) captures how well the integration maintains the clustering structure:
   \[
   ARI_\text{cell} = \frac{\text{Index observed} - \text{Index expected}}{\text{Max index} - \text{Index expected}}.
   \]
   Here, the observed index measures the agreement between clustering results and the ground truth, while the expected index accounts for random chance. We interpret \( ARI_\text{cell} \) scores ranging from 0 (random labeling) to 1 (perfect agreement).

\textbf{2. Normalized Mutual Information (\( NMI_\text{cell} \)):  }
   To evaluate how much information is shared between the true biological labels (\( Y \)) and the predicted cluster labels (\( C \)), we compute \( NMI_\text{cell} \) using:
   \[
   NMI_\text{cell} = \frac{2 \cdot I(Y; C)}{H(Y) + H(C)},
   \]
   where \( I(Y; C) \) is the mutual information, and \( H(Y) \) and \( H(C) \) are the entropies of the true and predicted labels. This score ranges from 0 (no alignment) to 1 (perfect alignment), and we use it to assess the consistency of clustering.

\textbf{3. Average Silhouette Width (\( ASW_\text{cell} \)):  }
   We calculate \( ASW_\text{cell} \) to measure how well-separated clusters are based on cell type labels. The silhouette score (\( ASW_C \)) evaluates whether cells are closer to their own cluster than to other clusters. We normalize the silhouette score using:
   \[
   ASW_\text{cell} = \frac{ASW_C + 1}{2}.
   \]
   This normalization ensures that the score ranges between 0 (poor separation) and 1 (perfect separation).

\subsection*{AvgBatch}

To assess how well the integration removes batch effects, we calculate \textbf{AvgBatch} as the average of two metrics: the Average Silhouette Width for batch labels (\( ASW_\text{batch} \)) and Graph Connectivity (\( GraphConn \)):

\[
\text{AvgBatch} = \frac{ASW_\text{batch} + GraphConn}{2}.
\]

\textbf{1. Average Silhouette Width for Batch Labels (\( ASW_\text{batch} \)):  }
   To evaluate the extent of batch effect removal, we compute \( ASW_\text{batch} \). First, we calculate the silhouette score based on batch labels (\( ASW_B \)), which measures how batch-specific artifacts affect the integrated space. Then, we adjust it to reflect batch mixing:
   \[
   ASW_\text{batch} = 1 - |ASW_B|.
   \]
   A higher \( ASW_\text{batch} \) score indicates better mixing of batches in the latent space, as cells are distributed independently of their batch origin.

\textbf{2. Graph Connectivity (\( GraphConn \)):  }
   To measure the connectivity of cells within the same biological type, we construct a k-nearest neighbors (kNN) graph for each cell type. Then, we identify the largest connected component (\( LCC \)) within each graph and calculate the connectivity score as:
   \[
   GraphConn = \frac{1}{|C|} \sum_{c \in C} \frac{|LCC(G_c^\text{kNN})|}{N_c}.
   \]
   Here, \( C \) is the set of cell types, \( |LCC(G_c^\text{kNN})| \) is the size of the largest connected component for cell type \( c \), and \( N_c \) is the total number of cells in \( c \). We use this score to evaluate whether cells of the same type remain well-connected after integration.

In the reconstruct experiments, we use \textbf{Exact Match}, \textbf{Levenshtein Distance}, and \textbf{BLEU Score} to evaluate the reconstruction ability of the GeneMamba model. We refer to the input gene token sequence as \( S_{\text{in}} \) and the model output sequence as \( S_{\text{out}} \). These metrics are defined as follows:

\subsection*{Exact Match}
The Exact Match (EM) measures whether the input sequence \( S_{\text{in}} \) is identical to the output sequence \( S_{\text{out}} \). It is defined as:

\[
\text{EM} = 
\begin{cases} 
1, & \text{if } S_{\text{in}} = S_{\text{out}}, \\
0, & \text{otherwise}.
\end{cases}
\]

For a dataset with \( N \) cells, the overall Exact Match score is the average:

\[
\text{EM}_{\text{avg}} = \frac{1}{N} \sum_{i=1}^N \text{EM}(S_{\text{in}}^i, S_{\text{out}}^i).
\]

\subsection*{Levenshtein Distance}
The Levenshtein Distance (LD) quantifies the minimum number of single-character edits (insertions, deletions, or substitutions) required to transform \( S_{\text{in}} \) into \( S_{\text{out}} \). Formally:

\[
\text{LD}(S_{\text{in}}, S_{\text{out}}) = \min 
\begin{cases} 
\text{LD}(S_{\text{in}}[:i-1], S_{\text{out}}[:j]) + 1, & \text{deletion}, \\
\text{LD}(S_{\text{in}}[:i], S_{\text{out}}[:j-1]) + 1, & \text{insertion}, \\
\text{LD}(S_{\text{in}}[:i-1], S_{\text{out}}[:j-1]) + c, & \text{substitution},
\end{cases}
\]

where \( c = 0 \) if \( S_{\text{in}}[i] = S_{\text{out}}[j] \), otherwise \( c = 1 \).

For normalization across sequences of varying lengths, the Normalized Levenshtein Distance is computed as:

\[
\text{NLD} = 1 - \frac{\text{LD}(S_{\text{in}}, S_{\text{out}})}{\max(|S_{\text{in}}|, |S_{\text{out}}|)},
\]

where \( |S| \) denotes the length of sequence \( S \).

\subsection*{BLEU Score}
The BLEU Score (Bilingual Evaluation Understudy) evaluates the similarity between \( S_{\text{out}} \) and \( S_{\text{in}} \) by comparing n-grams. It is computed as:

\[
\text{BLEU} = \exp \left( \sum_{n=1}^N w_n \log p_n \right) \cdot \text{BP},
\]

where:
\begin{itemize}
    \item \( p_n \) is the precision of \( n \)-grams,
    \item \( w_n \) is the weight for \( n \)-grams (often \( w_n = \frac{1}{N} \) for uniform weights),
    \item \( \text{BP} \) is the brevity penalty to penalize short outputs.
\end{itemize}

The brevity penalty is defined as:

\[
\text{BP} = 
\begin{cases} 
1, & \text{if } |S_{\text{out}}| \geq |S_{\text{in}}|, \\
\exp(1 - \frac{|S_{\text{in}}|}{|S_{\text{out}}|}), & \text{otherwise}.
\end{cases}
\]

The BLEU score ranges from 0 (no similarity) to 1 (perfect match).

These metrics together provide a comprehensive assessment of the GeneMamba model’s ability to reconstruct input gene token sequences accurately.

\subsection*{Baselines}
\label{section:baselines}

To evaluate the performance of our proposed approach, we compare it with several state-of-the-art baseline methods. Below, we briefly describe each method:

\begin{itemize}
    \item \textbf{GeneFormer}~\cite{theodoris2023transfer}: A transformer-based model specifically designed for single-cell analysis. It uses a gene-token representation to capture complex gene-gene relationships and downstream cellular heterogeneity.

    \item \textbf{scGPT}~\cite{cui2024scgpt}: A generative pre-trained transformer adapted for single-cell data analysis. scGPT utilizes a pretraining-finetuning paradigm to handle various single-cell tasks, including cell-type classification and clustering.

    \item \textbf{scFoundation}~\cite{hao2024large}: A foundational model for single-cell data analysis that incorporates self-supervised learning techniques to generate robust embeddings for single-cell profiles. It excels in tasks requiring integration across batches and datasets.

    \item \textbf{scBERT}~\cite{yang2022scbert}: A BERT-like architecture adapted for single-cell data. scBERT leverages bidirectional contextual embeddings to model complex interactions between genes within cells, making it suitable for both classification and clustering tasks.

    \item \textbf{Harmony}~\cite{korsunsky2019fast}: A batch-effect correction method that aligns single-cell datasets from different conditions or experiments. This method is often used as a preprocessing step to ensure integration and compatibility across datasets before applying downstream machine learning methods.
\end{itemize}

\subsection*{Finetuning Details}

For the cell type annotation task, we append the [CLS] token and add a multi-layer perceptron (MLP) to further train the model on the downstream task in a supervised manner. The train/test split is predefined by the dataset provider, so we do not perform any additional splitting. We use the generated [CLS] token embedding as the representation of the cell.

For the multi-batch integration task, we utilize the model fine-tuned on the cell type annotation task, remove the MLP layer, and directly extract embeddings from the GeneMamba model to conduct the experiment.

\subsection*{Gene Analysis Experiment}

The gene embeddings for different models are obtained as follows:

\begin{itemize}
    \item \textbf{Gene2Vec:} The embeddings are generated using the Gene2Vec model, where each gene corresponds to a unique embedding.
    \item \textbf{scGPT:} The embeddings are extracted using the provided function of scGPT, with each gene mapped to a distinct embedding.
    \item \textbf{scFoundation:} Since the scFoundation model does not provide a direct method for obtaining gene embeddings, we apply mean pooling to the generated context-aware embeddings of scFoundation on downstream datasets.
\end{itemize}

\section{Gene Correlation Analysis}
\label{section: Gene Correlation Analysis}
\begin{figure}[ht]
\begin{center}
\centerline{\includegraphics[width=\columnwidth]{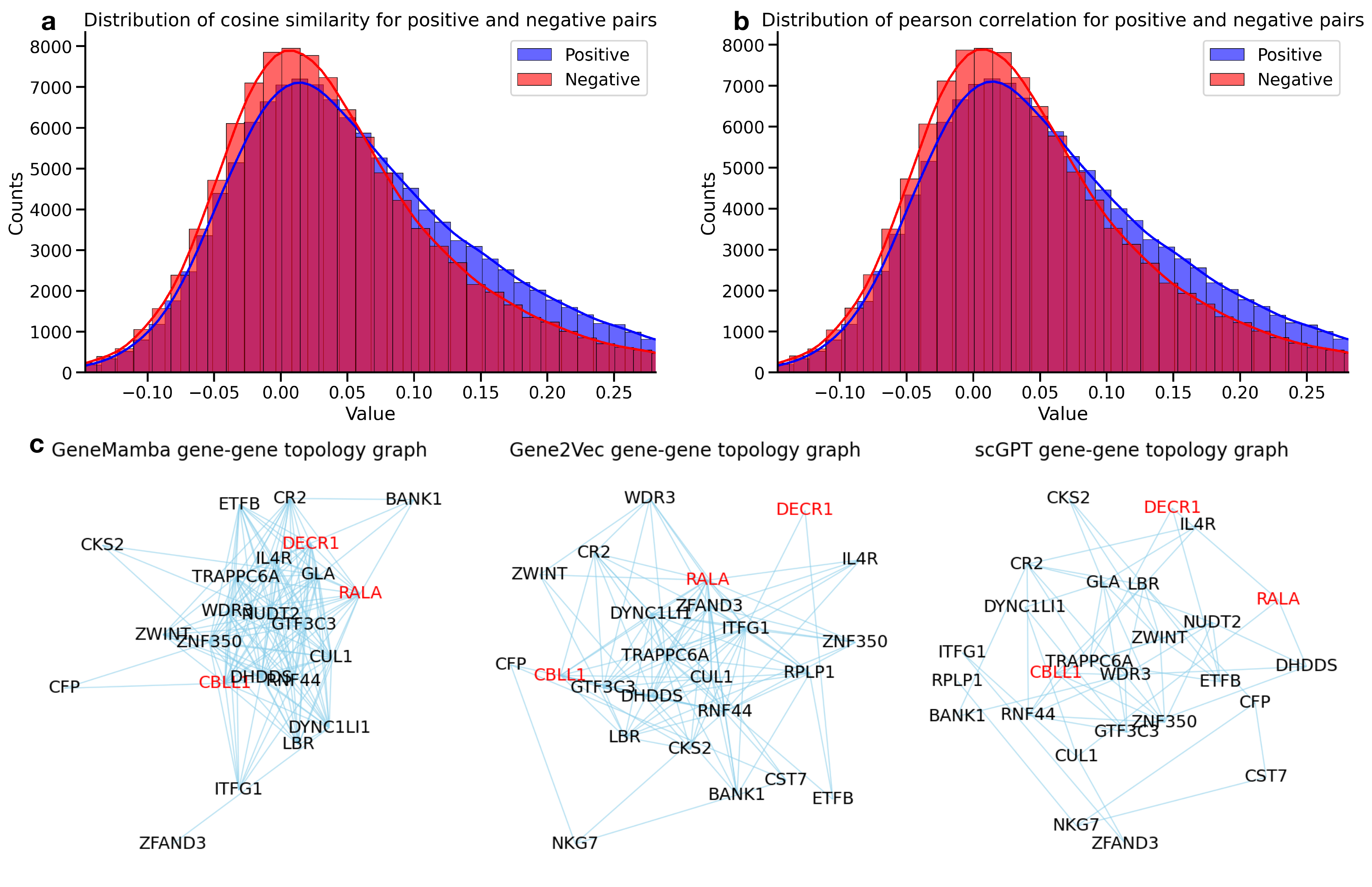}}
\caption{\textbf{Gene Correlation Analysis.}  
(a) Distribution of consine similarity score for positive and negative gene pairs using Gene2Vec embeddings.
(b) Distribution of pearson correlation score for positive and negative gene pairs using Gene2Vec embeddings.
(c) Gene-gene topology comparison using adjacency matrices highlights shared and unique structures captured by GeneMamba, scGPT, and Gene2Vec, such as genes RALA, DECR1, and CBL1.
}
\label{fig:gene_correlation_analysis}
\end{center}
\end{figure}
GeneMamba exhibits exceptional capabilities in gene correlation analysis, distinguishing itself in capturing biologically meaningful patterns, maintaining raw data fidelity, and identifying unique model-specific topologies. A series of experiments evaluated GeneMamba’s performance in distinguishing gene relationships, preserving correlation structures, and classifying gene types. 

\textbf{Differentiating Positive and Negative Gene Pairs}
In the first experiment, GeneMamba’s ability to differentiate between "positive" and "negative" gene pairs was assessed using Gene2Vec embeddings. Nearly 30,000 gene pairs, evenly distributed as positive (1) or negative (0), were analyzed for similarity scores using cosine similarity and Pearson correlation. GeneMamba demonstrated a significant separation between the mean similarity scores of positive and negative pairs (Figure~\ref{fig:gene_correlation_analysis} a, b). Compared to baseline models, GeneMamba achieved a clearer distinction, emphasizing its capacity to accurately represent gene relationships.

\textbf{Embedding Alignment}
To quantify discrimination power, we compared GeneMamba’s embeddings with those of scGPT and scFoundation. The "distance" between positive and negative pair distributions was measured using Euclidean distance, KL divergence, and JS divergence. The result is shown in Appendix Figure~\ref{fig:appendix_gene-gene pairs correlation heatmap}. GeneMamba consistently outperformed the other models across all metrics, highlighting its superior ability to distinguish between gene pair types and reinforcing its robustness in modeling gene relationships.

\textbf{Gene-Gene Topology Analysis}  
The ability to capture gene-gene topologies was analyzed using adjacency matrices derived from embeddings. For the PBMC12K dataset, distances between embeddings of randomly selected gene pairs (100 cells in our case) were calculated and binarized based on threshold value. Then we build the topology graph based on the adjacent matrix (Figure ~\ref{fig:gene_correlation_analysis}c) for each evaluated model.  Notably, while identifying unique model-specific topologies, all three models captured biologically significant genes and their interactions, such as RALA, DECR1, and CBL1. This suggests that the foundation models share a similar representation space.




\section{Perturbation Analysis}

To more comprehensively evaluate the generalizability and biological relevance of GeneMamba, we conducted a perturbation prediction study using protocols established by scGPT \cite{cui2024scgpt}. Perturbation studies are critical for assessing a model's ability to simulate cellular responses to genetic interventions—a fundamental requirement in functional genomics and drug discovery. We employed the Perturb-seq dataset from Norman \cite{norman2019exploring}, which include two-gene combinatorial perturbations. Specifically, we compared Mamba-based GEARS embeddings with the original GEARS model (baseline), following the benchmarking criteria previously proposed.

\begin{figure}
    \centering
    \includegraphics[width=1.0\linewidth]{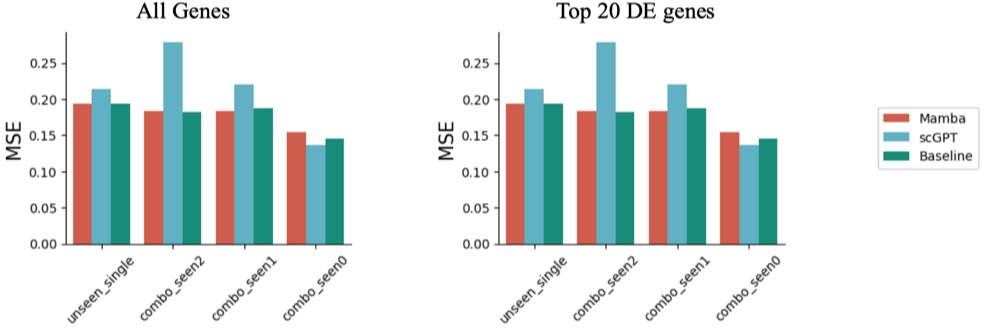}
    \caption{Mean Squared Error of DE analysis }
    \label{fig:de analysis}
\end{figure}

Since single-cell-level ground truth was unavailable, we evaluated performance using the averaged Mean Squared Error (MSE) of the top 20 differentially expressed (DE) genes before and after perturbation, and the result is shown in Figure \ref{fig:de analysis}. GeneMamba achieved lower MSE across these genes compared to the baseline, suggesting improved fidelity in modeling transcriptional changes. Additionally, we quantified prediction accuracy by measuring the proportion of top 20 DE genes whose predicted expression fell within the 45–55\% quantile range of the true expression distribution, where the Mamba-based model again showed superior alignment with ground truth.

\begin{figure}
    \centering
    \includegraphics[width=1.0\linewidth]{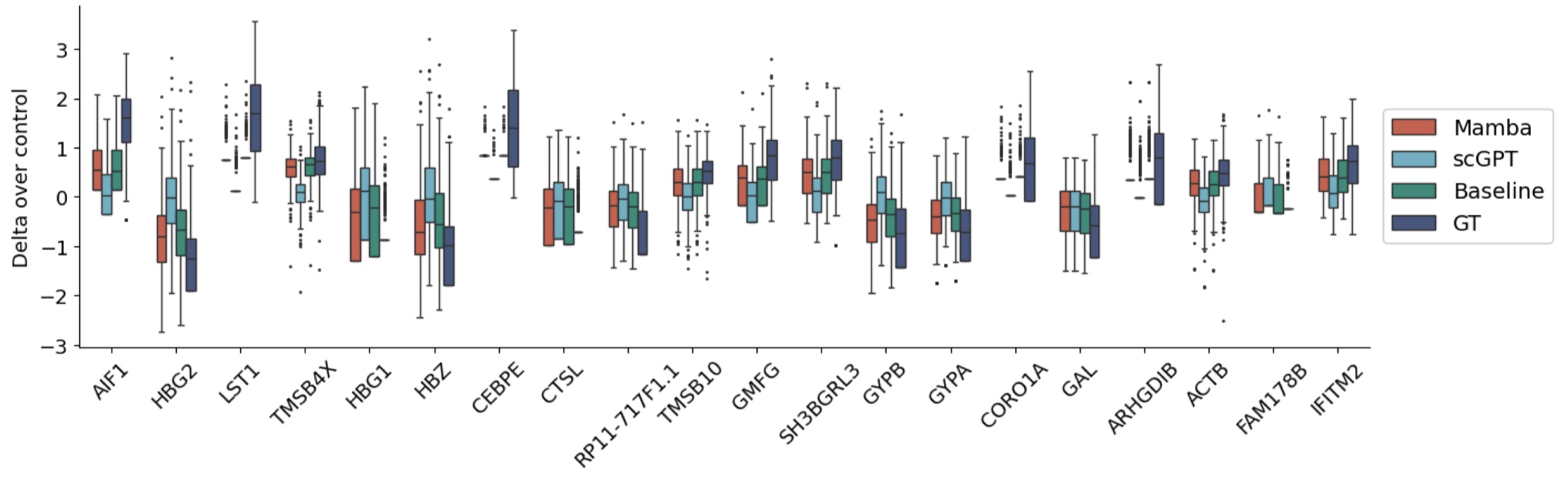}
    \caption{ETS2 + CEBPE perturbation result}
    \label{fig:ets2}
\end{figure}

\begin{figure}
    \centering
    \includegraphics[width=1.0\linewidth]{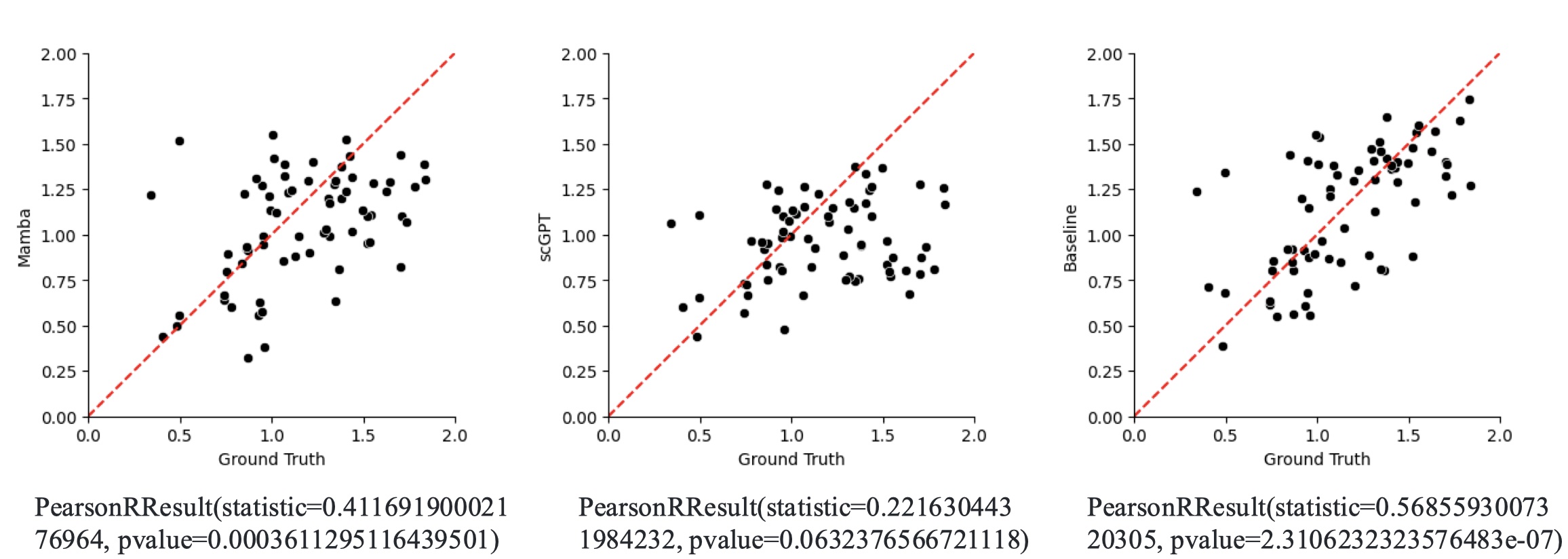}
    \caption{Magnitude scores computed for all test perturbing combinations on the Norman dataset.}
    \label{fig:magnitude}
\end{figure}

Furthermore, in a representative combinatorial perturbation case (ETS2 + CEBPE), as shown in Figure \ref{fig:ets2},  GeneMamba-generated predictions (red) closely matched the ground truth (purple) and outperformed the baseline (blue) in capturing post-perturbation expression shifts. Correlation analysis between predicted and ground-truth magnitude scores across all test perturbations yielded a higher Pearson coefficient (r = 0.5686, p < 1e-6) for GeneMamba, compared to the baseline GEARS (r = 0.4117), as shown in Figure \ref{fig:magnitude}. Lastly, the identification of suppressor and synergistic gene interactions revealed broader and more accurate overlap with the verified perturbation set using GeneMamba. These findings underscore the model’s enhanced interpretability and robustness under gene perturbation scenarios.

\section{More Related Work}
\label{section:more_related_work}

\subsection{Discretization of Gene Expression}

The transformation of gene expression levels into machine-readable tokens remains a critical aspect of single-cell modeling, with emerging methods aiming to balance biological relevance and computational efficiency.

Bin\-Based Strategy: Binning is the most commonly employed strategy for discretization, where continuous gene expression values are converted into categorical tokens, allowing transformers to process the data. Two main binning strategies are widely used:

Fixed-Size Binning: As seen in scBERT, fixed-size binning divides the range of expression values into equally sized intervals. Genes falling within the same interval are represented identically. This method, however, often leads to significant information loss, as most genes with low expression levels cluster within the same bin (e.g., the [0,1) bin \cite{boiarsky2023deep}), masking subtle but biologically relevant differences.

Adaptive Binning: Adaptive binning, employed by scGPT, dynamically adjusts bin sizes such that each interval represents an equal proportion of expressed genes within a single cell. While this approach aims to better capture the distribution of gene expression, it still faces challenges related to information loss, particularly for genes with extreme expression values or biological significance.

Value Projection Strategy: The value projection strategy, first introduced by TOSICA \cite{chen2023transformer}, has since been adopted by scCLIP \cite{xiong2023scclip}, SpaFormer \cite{wen2023single}, CellLM \cite{zhao2023large}, and CellPLM \cite{wen2023cellplm}. In this approach, continuous gene expression values are directly projected into high-dimensional embeddings through linear or non-linear transformations. This method retains the full resolution of the data, avoiding the information loss associated with binning. However, feeding continuous values into transformer-based models diverges from traditional tokenization approaches in NLP, potentially impacting model performance. Additionally, continuous embeddings may be more sensitive to noise and batch effects, necessitating robust preprocessing steps.

Rank-Based Strategy: The rank-based strategy was first introduced in iSEEEK \cite{shen2022universal} for single-cell transformers. This approach ranks gene expression values within each cell and retains only the top-ranked genes. The remaining genes are either truncated or represented as a lower-priority group. By focusing on the top-ranked genes, this strategy reduces sequence length, significantly improving computational efficiency. It also aligns well with the biological prioritization of highly expressed genes in specific cell states or conditions.

\subsection{Model Architectures for Single-Cell Analysis}

Recent advancements in model architectures have addressed the computational challenges of transformer-based models in single-cell RNA sequencing (scRNA-seq) analysis, with notable innovations including scTCA~\cite{scTCA}, a hybrid Transformer-CNN architecture designed for imputation and denoising of single-cell read counts to enhance downstream analyses; scHyena~\cite{scHyena}, a foundation model for full-length scRNA-seq analysis in brain tissues that uses a novel transformer architecture to process data without information loss, improving tasks like cell type classification and data imputation; xTrimoGene~\cite{gong2024xtrimogene}, an efficient and scalable representation learner employing an asymmetric encoder-decoder transformer architecture to reduce computational requirements while maintaining high accuracy in tasks such as cell type annotation and drug combination prediction; scTransSort~\cite{scTransSort}, a transformer-based model for intelligent cell type annotation that leverages self-attention mechanisms to extract features from scRNA-seq data, achieving high accuracy and performance; scGraphformer~\cite{scGraphformer}, which employs transformer-based graph neural networks to provide accurate and scalable cell type annotations, uncovering cellular heterogeneity and interactions in scRNA-seq data; scGFT~\cite{scGFT}, a train-free, cell-centric generative model adept at synthesizing scRNA-seq data to enhance data augmentation and analysis; White-Box Diffusion Transformer\cite{WhiteBoxDiffusionTransformer}, a hybrid model combining diffusion models and white-box transformers to generate synthetic and biologically plausible scRNA-seq data, improving training efficiency and resource utilization; STGRNS~\cite{STGRNS}, an interpretable transformer-based method for inferring gene regulatory networks from scRNA-seq data to enhance understanding of gene interactions; scRDiT~\cite{scRDiT}, a generative approach using diffusion transformers to create virtual scRNA-seq data, facilitating the study of gene expression dynamics; and TransformerST~\cite{TransformerST}, an unsupervised model based on the transformer architecture for super-resolution in spatial transcriptomics, improving spatial gene expression analysis. These models collectively advance the field by addressing key challenges in scRNA-seq data processing and interpretation.

Traditional single-cell large language models (LLMs) are predominantly based on transformer architectures~\cite{boiarsky2023deep, white2024comprehensive}. While transformers have demonstrated exceptional performance in various single-cell transcriptomics tasks, their high computational and memory requirements make training large models time-intensive and resource-heavy. These limitations become particularly evident when processing the ultra-long sequences typical of single-cell RNA-seq data. To alleviate these challenges, some transformer-based models have incorporated optimizations to improve efficiency. For example, scBERT integrates Performer \cite{choromanski2020rethinking} to approximate attention calculations, while scGPT employs Flash Attention \cite{dao2022flashattention} to accelerate training by reducing memory overhead. Despite these advancements, the overall time required to train such models remains substantial, especially for datasets with millions of cells and thousands of genes. To address the intrinsic limitations of transformers, Bidirectional Mamba (Bi-Mamba) has been proposed as a novel alternative architecture. Bi-Mamba leverages state-space models (SSMs) to process ultra-long sequences with linear computational complexity in both time and memory. This represents a significant departure from traditional transformer-based methods, offering a scalable and efficient solution for the high-dimensional, high-throughput nature of single-cell transcriptomics data.

\section{Discussion of Scalability}
\label{section:scalability}
Scalability is an essential feature for models designed to handle large-scale single-cell omics datasets. The GeneMamba model demonstrates superior scalability and computational efficiency across various configurations, including sequence lengths and model complexities. In this section, we discuss its scalability based on the results summarized in Tables~\ref{table:training_configurations} and~\ref{table:efficiency_comparison}.

\subsection*{Training Configurations and Model Complexity}
Table~\ref{table:training_configurations} presents the training configurations for the Mamba and GeneMamba models, highlighting the relationship between model complexity (e.g., number of layers, hidden size) and training requirements. For instance, increasing the number of layers from 24 to 48 doubles the parameters and correspondingly increases training time two times (from 5.5 hours to 11 hours). Additionally, GeneMamba demonstrates its capability to handle longer sequence lengths (e.g., 4096 tokens) with a linear increase in training time (from 5.5 hours to 11 hours), showcasing its scalability for more input gene tokens of single-cell data.

\begin{table}[ht]
\centering
\caption{Training configurations for the GeneMamba models, demonstrating scalability across layers, hidden sizes, and sequence lengths. The data highlights the increasing number of parameters and training time as model complexity grows, with GeneMamba showing efficient handling of longer sequence lengths (e.g., 4096) compared to traditional Mamba setups.}
\begin{tabular}{ccccc}
\toprule
\textbf{Mamba Layers} & \textbf{Hidden Size} & \textbf{Seq Length} & \textbf{Param Size} & \textbf{Training Time (h/million)} \\
\midrule
24 & 512 & 2048 & 65.74M & 5.5 \\
48 & 512 & 2048 & 105.42M & 10.5 \\
24 & 768 & 2048 & 127.05M & 11 \\
48 & 768 & 2048 & 215.03M & 22 \\
24 & 512 & 4096 & 65.74M & 11 \\
\bottomrule
\end{tabular}

\label{table:training_configurations}
\end{table}

\subsection*{Computational Efficiency Comparison}
Table~\ref{table:efficiency_comparison} compares the computational efficiency of the Transformer and GeneMamba backbones. GeneMamba requires significantly fewer FLOPs per sample and achieves faster training times across all sequence lengths. For example, at a sequence length of 2048, GeneMamba's training time is 0.1284 seconds/sample compared to the Transformer's 0.2227 seconds/sample, a reduction of approximately 42\%. This efficiency becomes even more pronounced at longer sequence lengths.

\begin{table}[ht]
\centering
\caption{Computational efficiency comparison of Transformer and GeneMamba backbones at different input sequence lengths. GeneMamba significantly reduces computational cost while maintaining high efficiency.}
\resizebox{\textwidth}{!}{%
\begin{tabular}{ccccc}
\toprule
\textbf{Model Backbone} & \textbf{Parameters} & \textbf{FLOPs/sample} & \textbf{Seq Length} & \textbf{Training Time (s/sample)} \\
\midrule
Transformer & 130M & 2.85E+12 & 2048 & 0.2227 \\
Transformer & 130M & 7.18E+12 & 4096 & 0.6025 \\
Transformer & 130M & 2.02E+13 & 8192 & 1.8904 \\
GeneMamba   & 130M & 1.58E+12 & 2048 & 0.1284 \\
GeneMamba   & 130M & 3.15E+12 & 4096 & 0.2393 \\
GeneMamba   & 130M & 6.31E+12 & 8192 & 0.4885 \\
\bottomrule
\end{tabular}%
}
\label{table:efficiency_comparison}
\end{table}

\subsection*{Theoretical Computation and Demonstration}
Using the 2048-sequence-length GeneMamba model as an example, the total FLOPs required for processing 2 million cells can be calculated as:
\[
\text{Total FLOPs required} = 1.58 \times 10^{12} \times 2 \times 10^{6} = 3.16 \times 10^{18} \text{ FLOPs.}
\]

Given that the A100 80GB GPU has a peak performance of 19.5 TFLOPS for single-precision (FP32) and 312 TFLOPS for half-precision (FP16), the theoretical training time is computed as follows:

For FP32:
\[
t = \frac{3.16 \times 10^{18} \text{ FLOPs}}{19.5 \times 10^{12} \text{ FLOPs/sec}} \approx 1.621 \times 10^{5} \text{ seconds} = 45 \text{ hours.}
\]

For FP16:
\[
t = \frac{3.16 \times 10^{18} \text{ FLOPs}}{312 \times 10^{12} \text{ FLOPs/sec}} \approx 1.013 \times 10^{4} \text{ seconds} = 2.8 \text{ hours.}
\]

These results demonstrate the substantial efficiency improvements provided by GeneMamba, especially when leveraging FP16 precision, which reduces training time from 45 hours to just 2.8 hours.

\subsection*{Conclusion on Scalability}
GeneMamba's ability to process longer sequences with lower computational cost and its effective utilization of GPU resources make it a scalable solution for large-scale single-cell analysis. As shown in Tables~\ref{table:training_configurations} and~\ref{table:efficiency_comparison}, its scalability and efficiency significantly outperform traditional Transformer architectures, providing a robust framework for computational biology and related applications.

\section{Extended Figures}
\label{section:appendix_more_figures}

\begin{figure}[H]
\begin{center}
\centerline{\includegraphics[width=0.6\columnwidth]{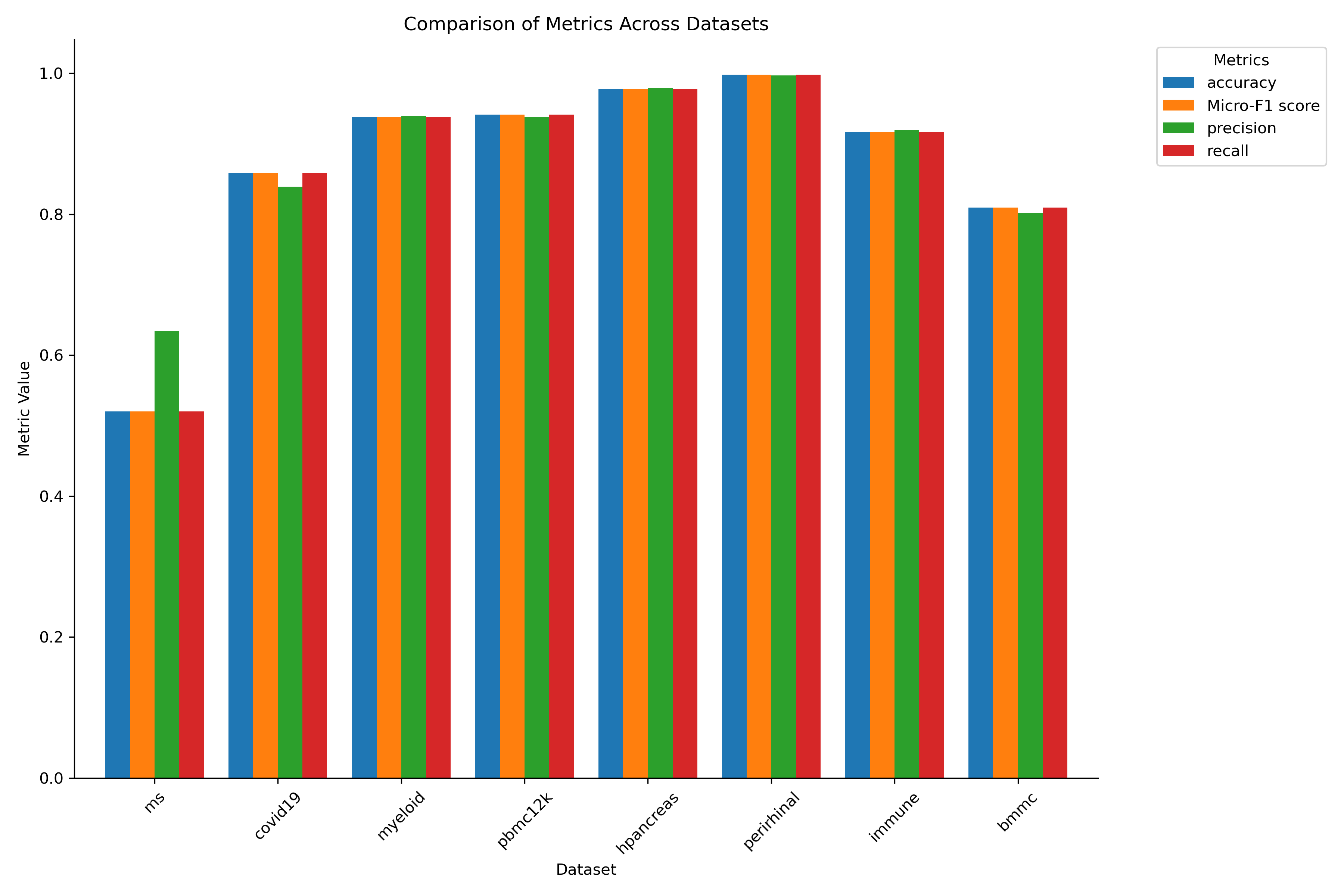}}
\caption{\textbf{Results of cell type annotation.} Bar plot of the classification metrics across various datasets of GeneMamba.}
\label{fig:appendix_barplot_classification_results}
\end{center}
\end{figure}

\begin{figure}[H]
\begin{center}
\centerline{\includegraphics[width=\columnwidth]{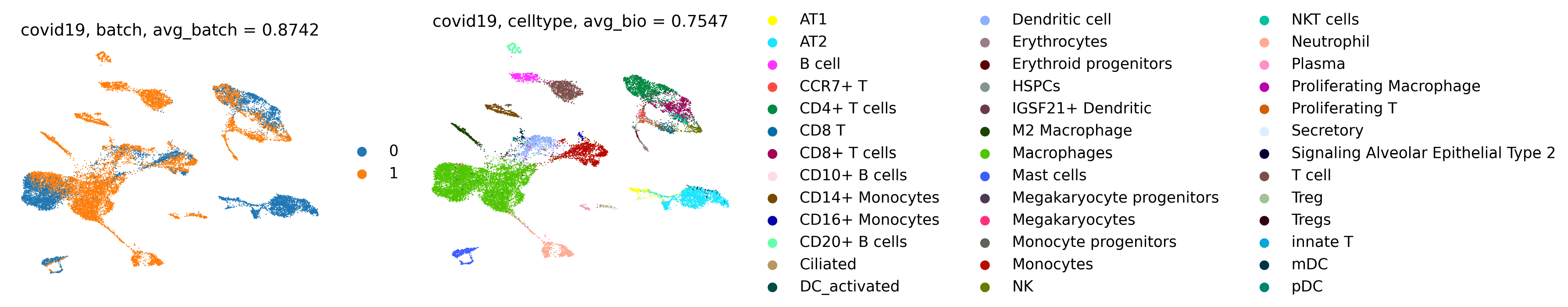}}
\caption{\textbf{Results of multi-batch integration.} Benchmark of the fine-tuned GeneMamba on the Covid19 dataset for the multi-batch integration task. The UMAP plot of learned cell embeddings is colored by cell types.}
\label{fig:appendix_multi_batch_covid19}
\end{center}
\end{figure}

\begin{figure}[H]
\begin{center}
\centerline{\includegraphics[width=\columnwidth]{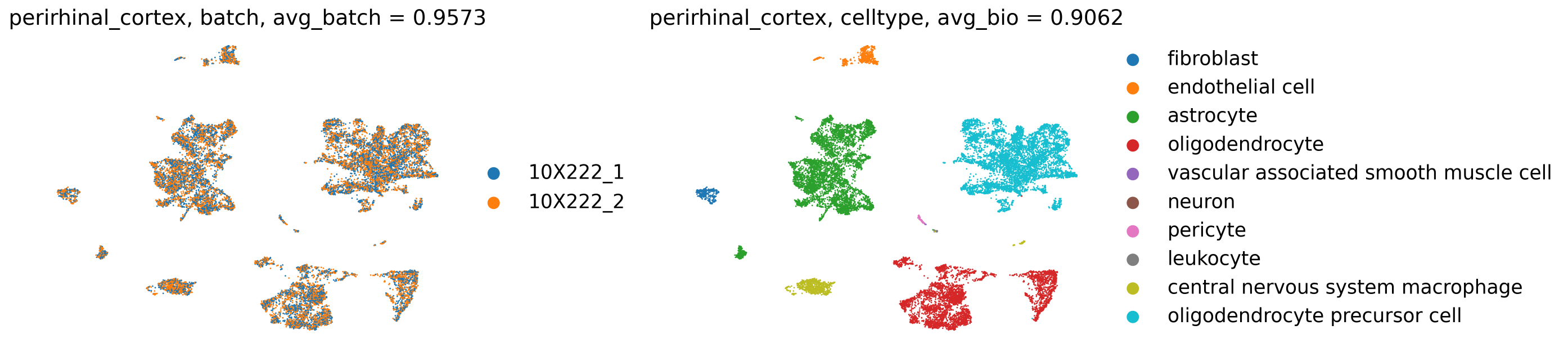}}
\caption{\textbf{Results of multi-batch integration.} Benchmark of the fine-tuned GeneMamba on the Perirhinal Cortex dataset for the multi-batch integration task. The UMAP plot of learned cell embeddings is colored by cell types.}
\label{fig:appendix_multi_batch_percor}
\end{center}
\end{figure}

\begin{figure}[H]
\begin{center}
\centerline{\includegraphics[width=0.8\columnwidth]{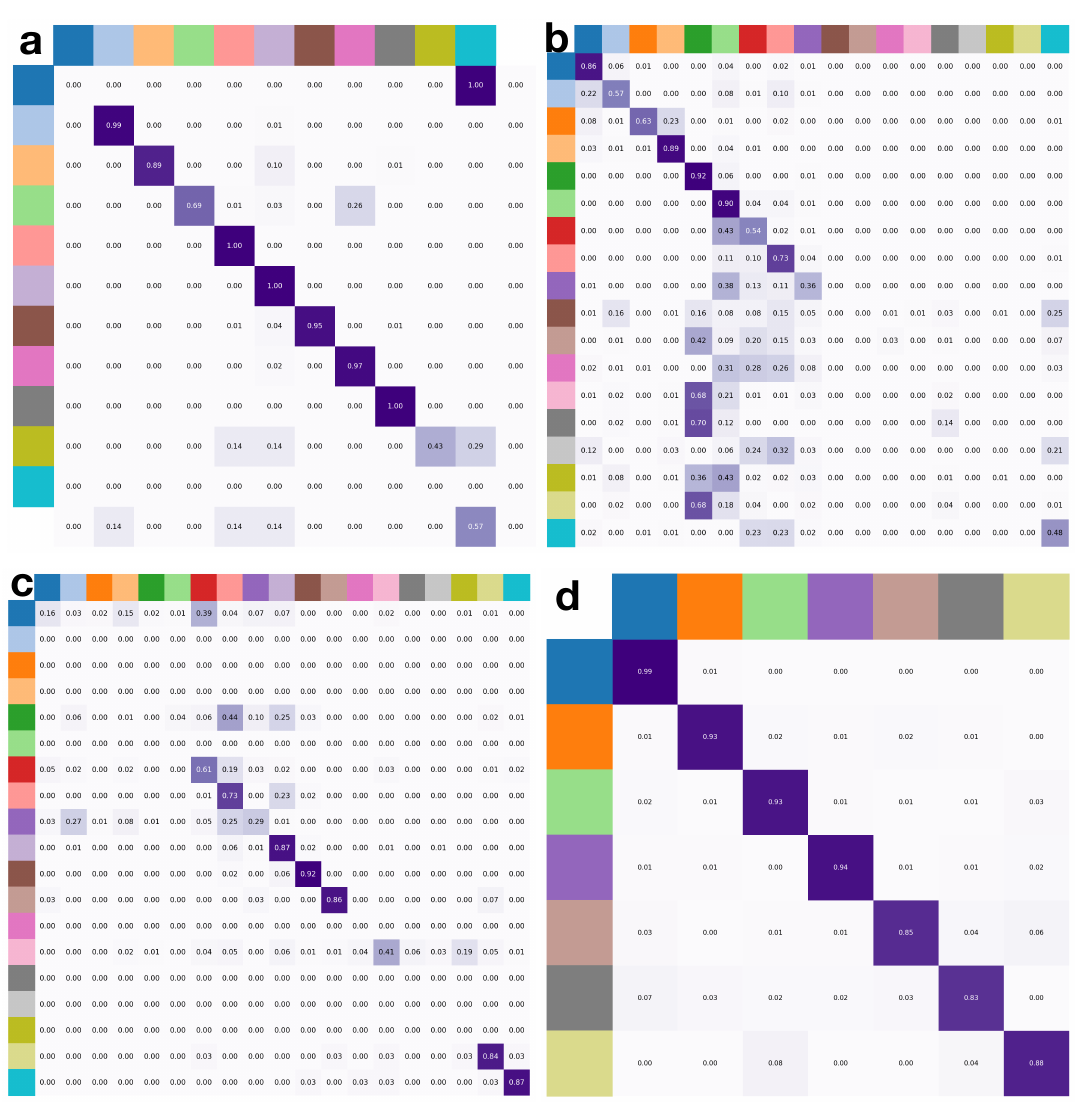}}
\caption{\textbf{Cell type annotation results of GeneMamba.} a. Confusion matrix of dataset hPancreas. b. Confusion matrix of dataset MS. c. Confusion matrix of dataset Myeloid. d. Confusion matrix of dataset Myeloid\_b.}
\label{fig:cm_four_datasets}
\end{center}
\end{figure}

\begin{figure}[H]
\begin{center}
\centerline{\includegraphics[width=0.8\columnwidth]{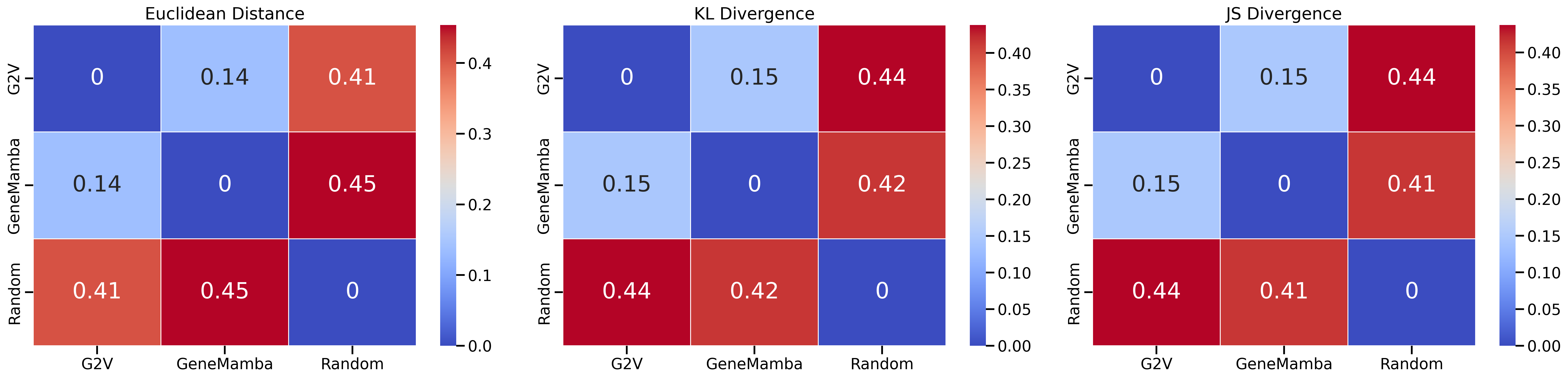}}
\caption{\textbf{Gene-gene pairs correlation analysis.} The Euclidean Distance, KL Divergence, and JS Divergence across three embedding methods (Gene2Vec, GeneMamba, and Random). A lower value indicates higher similarity. The embeddings generated by GeneMamba are more similar to Gene2Vec embeddings compared to randomly generated embeddings.}
\label{fig:appendix_gene-gene pairs correlation heatmap}
\end{center}
\end{figure}

\section{Extended Tables}

\begin{table}[H]
\centering
\caption{Gene Rank Reconstruction Performance comparison of GeneMamba, GeneFormer, and GeneMamba\_S models across datasets.}
\begin{tabular}{c c c c c}
\hline
\textbf{Datasets} & \textbf{Models} & \textbf{L-Dist} & \textbf{BLEU} & \textbf{Spearman} \\ \hline
\multirow{3}{*}{PBMC12k} & GeneMamba\_U    & 430             & 0.532         & 0.469             \\ 
                         & GeneFormer      & 23              & 0.968         & 0.703             \\ 
                         & GeneMamba       & \textbf{6}      & \textbf{0.987}& \textbf{0.711}    \\ \hline
\multirow{3}{*}{Pancreas} & GeneMamba\_U    & 370             & 0.524         & 0.461             \\ 
                         & GeneFormer      & 25              & 0.956         & 0.763             \\ 
                         & GeneMamba       & \textbf{12}     & \textbf{0.991}& \textbf{0.792}    \\ \hline
\multirow{3}{*}{Zheng68k} & GeneMamba\_U    & 432             & 0.581         & 0.503             \\ 
                         & GeneFormer      & 25              & 0.937         & 0.901             \\ 
                         & GeneMamba       & \textbf{11}     & \textbf{0.996}& \textbf{0.980}    \\ \hline
\multirow{3}{*}{Immune}   & GeneMamba\_U    & 468             & 0.659         & 0.442             \\ 
                         & GeneFormer      & 17              & 0.962         & 0.823             \\ 
                         & GeneMamba       & \textbf{12}     & \textbf{0.998}& \textbf{0.844}    \\ \hline
\end{tabular}
\label{tab:appendix_gene_rank_reconstruct_models_comparison}
\end{table}

\newpage

\end{document}